\documentclass[10pt,twocolumn,letterpaper]{article}

\usepackage{iccv}  
\usepackage{hyperref}
\usepackage{multirow}
\usepackage{pifont}
\usepackage[table,dvipsnames,xcdraw]{xcolor}

\begin{document}
\title{VoD-3DGS: View-opacity-Dependent 3D Gaussian Splatting}
\author{Mateusz Nowak \and Wojciech Jarosz \and Peter Chin \\
Dartmouth College \\
Hanover, New Hampshire, USA \\
{\tt\small \{mateusz.m.nowak.th,wojciech.k.jarosz,Peter.Chin\}@dartmouth.edu}
}
\maketitle
\begin{figure*}[th]
    \centering
    \begin{subfigure}{\textwidth}
        \begin{subfigure}{\linewidth}
            \centering
            \begin{subfigure}{\linewidth}
            \centering
            \begin{subfigure}{\linewidth}
                  \centering
                  \includegraphics[width=\linewidth]{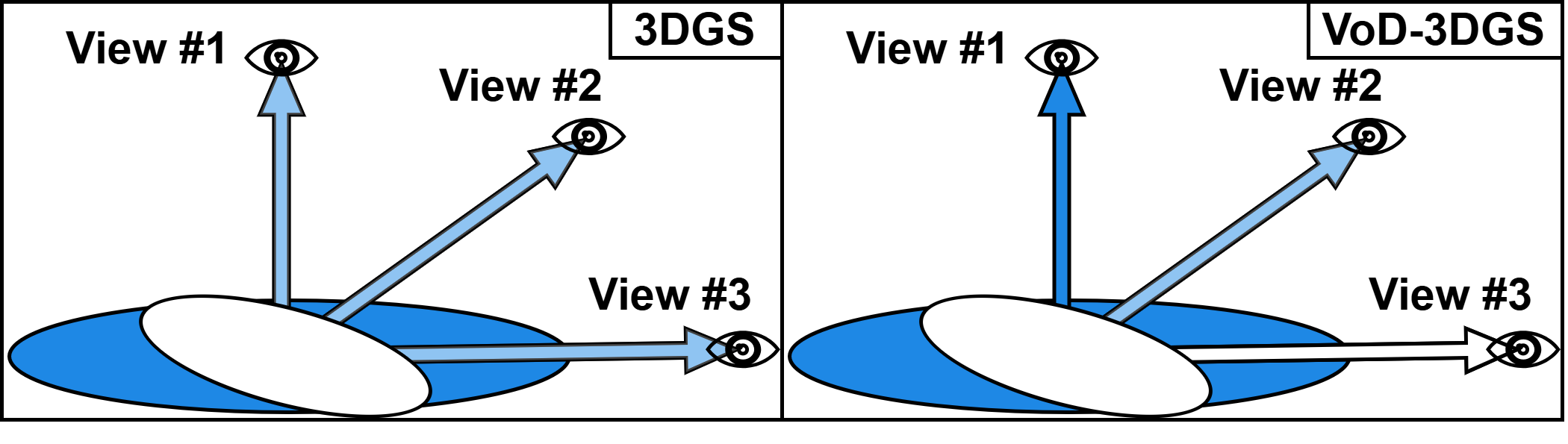}
            \end{subfigure}
            \end{subfigure}
            \begin{subfigure}{0.33\linewidth}
                \centering
                \includegraphics[width=\linewidth]{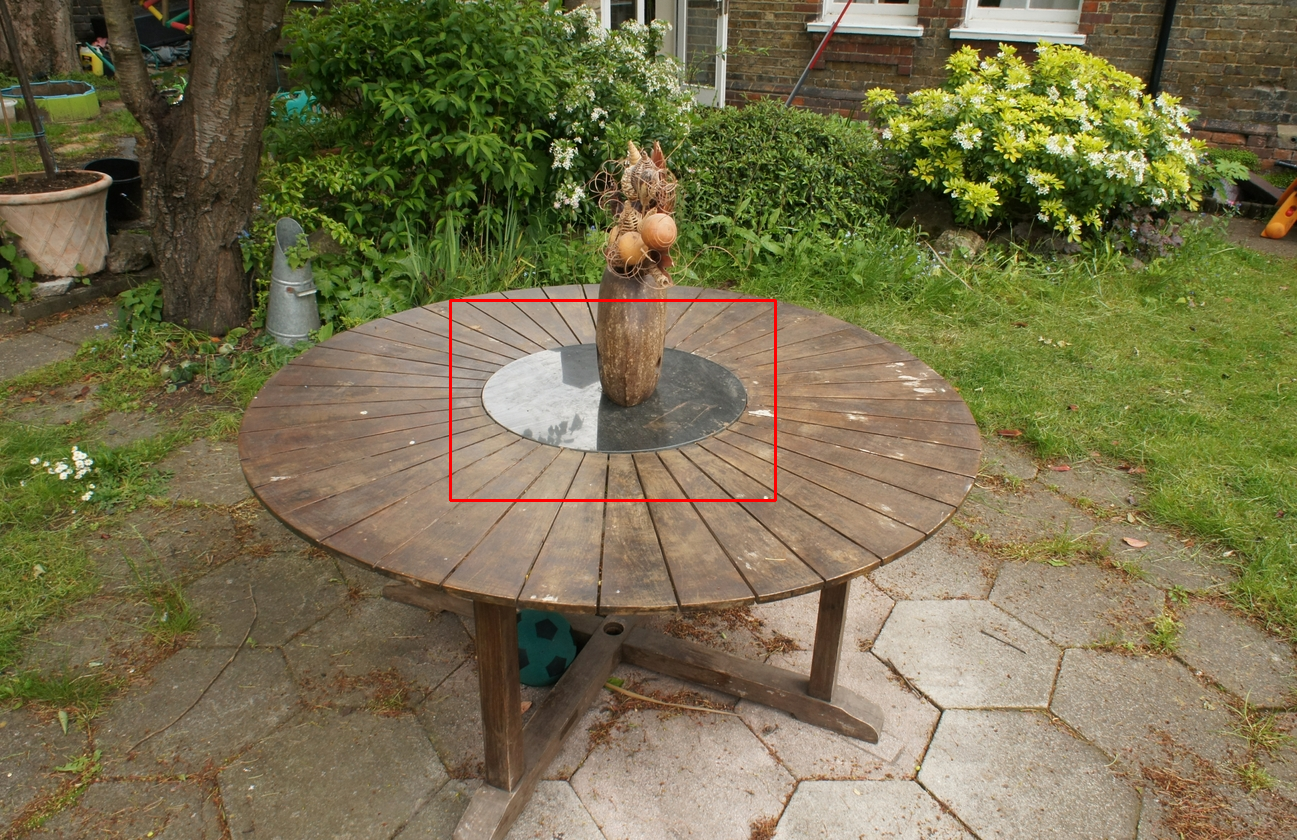}
                \subcaption{Ground Truth}   
            \end{subfigure}
            \begin{subfigure}{0.33\linewidth}
                  \centering
                  \includegraphics[width=\linewidth]{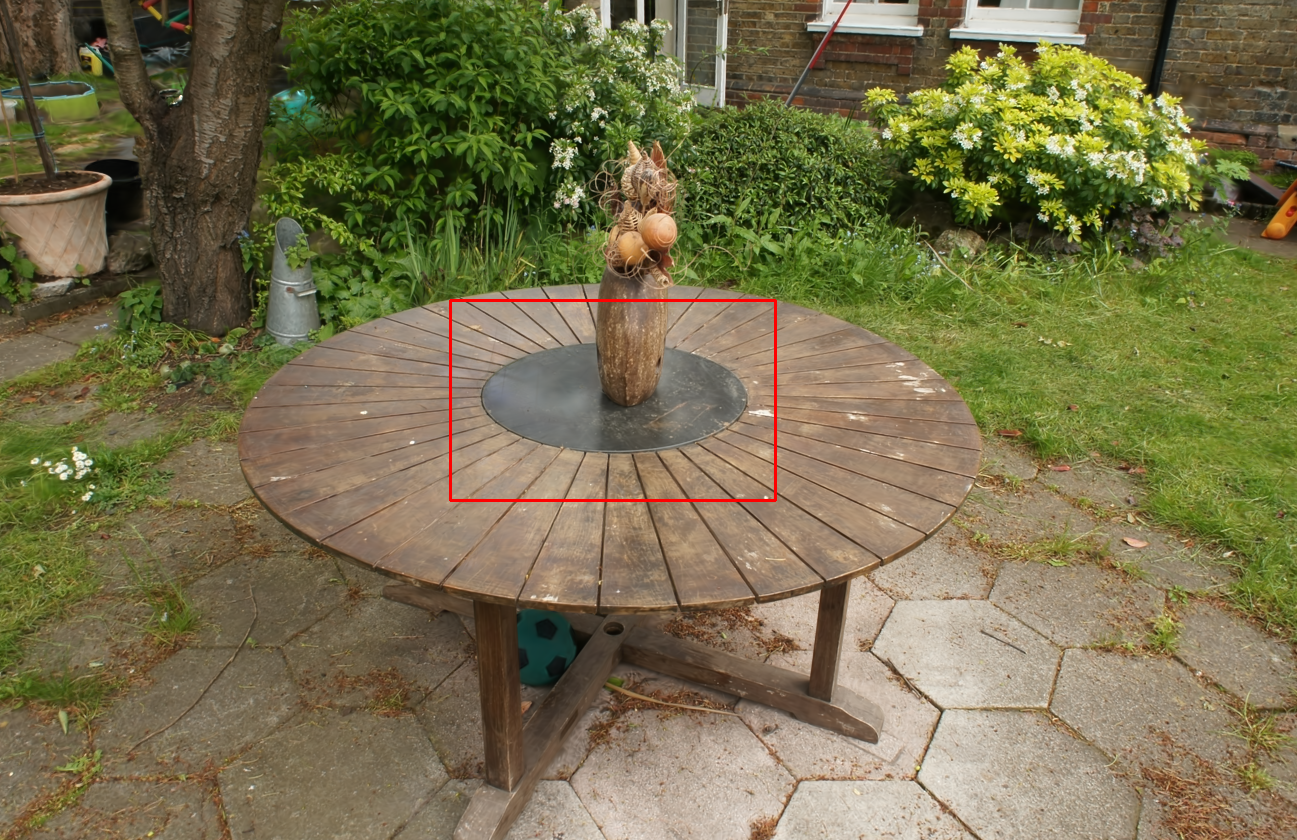}
                  \subcaption{3DGS} 
            \end{subfigure}
            \begin{subfigure}{0.33\linewidth}
                  \centering
                  \includegraphics[width=\linewidth]{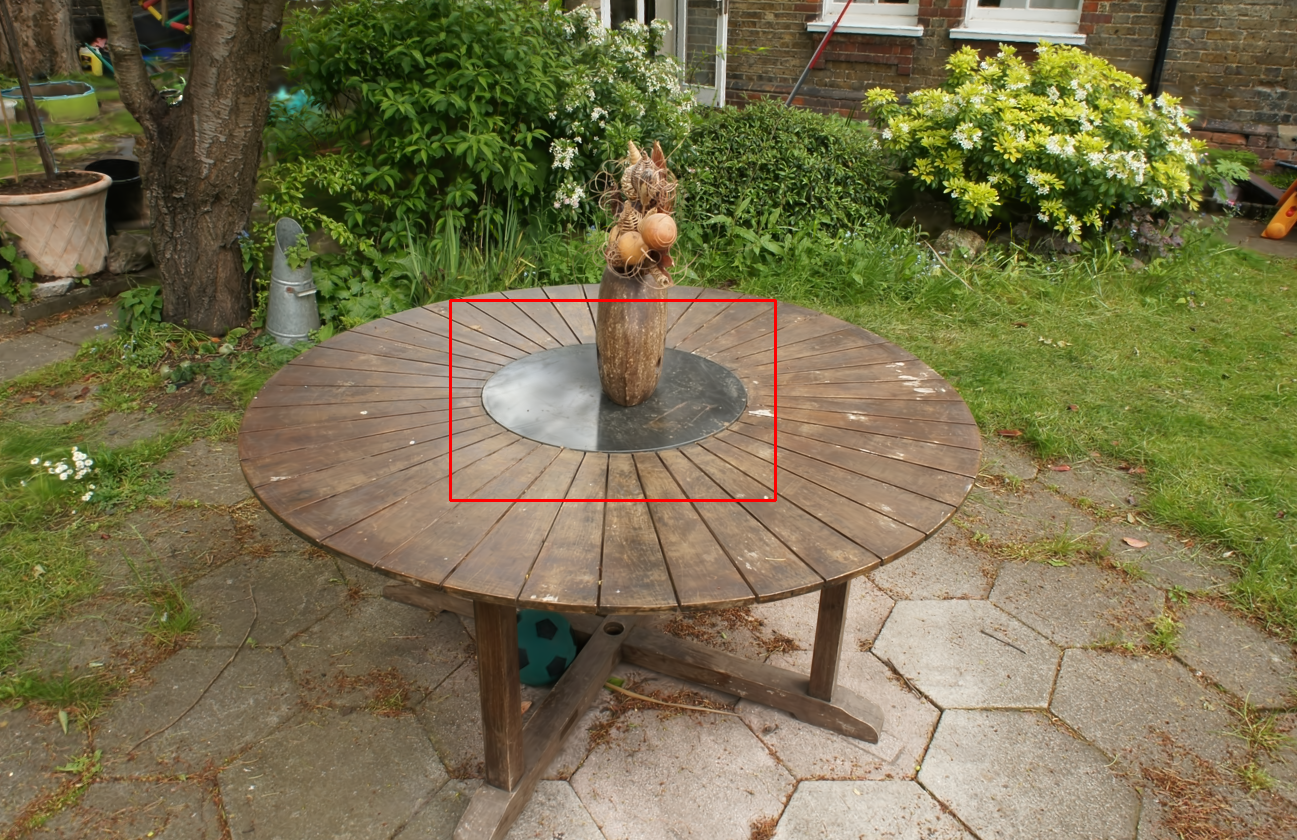}
                  \subcaption{VoD-3DGS (Our method)} 
            \end{subfigure}
        \end{subfigure}
    \end{subfigure} 
  \caption{[Top] The visualization of the effect of our method in comparison to the standard 3D Gaussian Splatting. We can suppress or boost the impact of Gaussians responsible for modeling specular highlights and reflections for certain views by extending the opacity component of each 3D Gaussian with a symmetric matrix, acting as a learnable view-dependent factor. [Bottom] Qualitative results of our proposed method in comparison to standard 3DGS. When multiplied with the view vector, the symmetric matrix allows us to represent reflections, specular highlights, and even changing lights.}
  \label{fig:teaser}
\end{figure*}

\begin{abstract}
Reconstructing a 3D scene from images is challenging due to the different ways light interacts with surfaces depending on the viewer's position and the surface's material. In classical computer graphics, materials can be classified as diffuse or specular, interacting with light differently. The standard 3D Gaussian Splatting model struggles to represent view-dependent content, since it cannot differentiate an object within the scene from the light interacting with its specular surfaces, which produce highlights or reflections. In this paper, we propose to extend the 3D Gaussian Splatting model by introducing an additional symmetric matrix to enhance the opacity representation of each 3D Gaussian. This improvement allows certain Gaussians to be suppressed based on the viewer's perspective, resulting in a more accurate representation of view-dependent reflections and specular highlights without compromising the scene's integrity. By allowing the opacity to be view dependent, our enhanced model achieves state-of-the-art performance on Mip-Nerf, Tanks\&Temples, Deep Blending, and Nerf-Synthetic datasets without a significant loss in rendering speed, achieving $>60FPS$, and only incurring a minimal increase in memory used.
\end{abstract}

\section{Introduction}
3D reconstruction describes recovering the 3D structure from a set of images for a particular scene. It is an essential task in computer vision; given a reconstructed 3D scene, the user can generate novel, unseen views, which can further areas like robotics, autonomous driving, content generation, and Augmented Reality. As an early advancement, Structure-from-Motion \cite{sfm} enabled a seamless transition from 2D images to 3D worlds. Nevertheless, these methods struggled with novel view synthesis and achieving a better understanding of complex scenes. With the rise of deep learning, NeRFs \cite{nerf} offered a direct mapping from 3D positions to colors and density, allowing a compact and continuous representation that enables novel view synthesis with photorealism and detail. However, even with significant improvements to the standard approach \cite{zipnerf, mipnerf360}, NeRFs still require a substantial amount of computing power to learn and render this 3D representation.

In recent years, 3D Gaussian Splatting \cite{3DGS} has offered a new approach to 3D reconstruction by providing a quick and efficient representation that allows for real-time rendering. As NeRF represents the scene through an implicit function, 3D Gaussian Splatting represents the scene explicitly through a collection of 3D anisotropic Gaussians, where each Gaussian is parameterized by its color, scale, rotation, opacity, and position in space. However, since 3D Gaussian Splatting uses rasterization to visualize the scene, it struggles with the photorealistic representation of various light interactions.

A plethora of solutions have appeared to extend the 3D Gaussian model through various improvements, from an improved shading function \cite{gs-shader}, novel shape representation \cite{2dgs, ges}, and ray tracing \cite{3dgrt} to the introduction of a machine learning model that enhances the capabilities of 3D Gaussians \cite{spec-gaussian}. Nevertheless, these models either struggle with efficient rendering and fail to provide a simple and compact 3D representation, or they do not enhance the photorealism of the scene regarding light interactions.

In this paper, we present a straightforward yet effective enhancement to standard 3D Gaussian Splatting. Specifically, we extend the traditional scalar opacity model to a view-dependent function, which allows for a more effective representation of specular highlights, dynamic lighting, and other light interactions without sacrificing rendering speed, extending the photorealism of the scene. We validate our method across multiple datasets and demonstrate state-of-the-art results that achieve $> 60$ FPS on complex real-life scenes.
\section{Related Work}
In recent years, novel view synthesis has seen notable advancement from implicit models like Neural Radiance Fields (NeRF) \cite{nerf}, where the scene is parametrized by a multi-layer perceptron (MLP) to represent geometry and view-dependent appearance, to explicit 3D Gaussian Splatting (3DGS) \cite{3DGS}, where geometry is represented by thousands or millions of anisotropic 3D Gaussians, whose appearances are directly stored in their representation via learnable per-Gaussian opacity and Spherical Harmonics coefficients.

\subsection{Neural Radiance Fields (NeRF)}
Both models have undergone further improvements, leading to reduced training times and improved quality. Instant-NGP \cite{instantNGP} introduced a novel feature-grid-based learnable scene representation that increases the expressive power while decreasing the training time of NeRFs. Models such as Mip-NeRF \cite{mipnerf360} and Zip-NeRF \cite{zipnerf} reduce aliasing artifacts by reasoning about sub-volumes along a cone. Additionally, to improve the visual quality of reflections and specular highlights, methods like Ref-NeRF \cite{refNerf} extend the NeRF model by reparameterizing the view-dependent outgoing radiance, showing the inherent lack of understanding of specular light interactions within the scene by standard NeRFs.

\subsection{Modifying the material model of 3DGS}

Various extensions have been proposed in the 3D Gaussian Splatting literature to modify the underlying material parameters. These extensions not only enhanced the photorealism of the scene but improved the representation of specular materials, better capturing the specular highlights and reflections. \Citet{gs-shader} provided a novel normal estimation with a simplified shading function on 3D Gaussians to enhance the neural rendering in scenes with reflective surfaces. \Citet{spec-gaussian} introduced a new 3D Gaussian representation that utilizes an anisotropic spherical Gaussian (ASG) appearance field with an MLP to model the appearance of each 3D Gaussian.  

We propose tackling the specular representation problem in the 3D Gaussian Splatting model without a machine learning model or an extended shading function that might negatively influence the rendering and training speed.

\subsection{Modifying the shape of 3DGS}
The Gaussian model was also enhanced in terms of its geometric representation. \Citet{2dgs} flattened the 3D Gaussian model into anisotropic 2D Gaussian discs to better capture the underlying 3D geometry surface. \Citet{ges} used the Generalized Exponential Function (GEF) to model 3D scenes instead of the standard 3D Gaussians. \Citet{gof} proposed to extract Gaussian Opacity Fields (GOF), by leveraging an explicit ray-Gaussian and defining the opacity of any 3D point as the minimal opacity among all training views that observed the point. Finally, \Citet{3dcs} utilized 3D smooth convexes as primitives for modeling radiance fields from multi-view images.

These methods show a simple paradigm in 3D Gaussian Splatting literature - we can achieve improved, more photorealistic results with simple introductions to the underlying 3DGS model. Similarly, we propose one of these simple modifications to extend the 3D Gaussian Splatting model with a view-dependent opacity.

\textit{In this paper, we extend the 3DGS model by introducing a learnable per-Gaussian symmetric matrix. This matrix transforms the scalar opacity into a function that varies with the viewing direction, allowing for improved representation of specular highlights and reflections without any neural network introduced. Notably, this enhancement does not result in any significant increase in rendering time or memory usage.}
\section{Preliminaries}

\subsection{3D Gaussian Splatting (3DGS)}

\Citet{3DGS} proposed parameterizing the scene with a set of learnable 3D Gaussian primitives, rendering images using differentiable rasterization through volume splatting. Each 3D Gaussian can be represented by its respective 3D covariance matrix $\boldsymbol{\Sigma}$ and the position of its center $\mu$:
\begin{equation}
    G(x) = e^{-\frac{1}{2}(x-\mu)^T\Sigma^{-1}(x-\mu)},
\end{equation}
where the covariance matrix $\Sigma = \mathbf{RS}\mathbf{S}^T\mathbf{R}^T$ is factorized into the corresponding learnable scaling matrix $\mathbf{S}$ and rotation matrix $\mathbf{R}$ to guarantee positive semi-definiteness of the matrix $\Sigma$. Then, the 3D Gaussian is transformed into camera coordinates, using a viewing transformation $\mathbf{W}$, and projected onto the image plane, via the Jacobian of the affine approximation of the projective
transformation $\mathbf{J}$ \cite{ewa}:

\begin{equation}
    \boldsymbol\Sigma' = \mathbf{JW}\boldsymbol\Sigma\mathbf{J}^T\mathbf{W}^T.
\end{equation}

Projecting each of the 3D Gaussians onto a 2D plane, we obtain a 2D Gaussian representation $G^{2D}$, which can be efficiently sorted and then rasterized using a volumetric $\alpha$ blending approach, where each color $C$ can be defined as:
\begin{equation}
    C(p) = \sum_{i = 1}^{N} \mathbf{c_i} \alpha_i G^{2D}_{i}(x) \prod_{j=1}^{i-1}(1-\alpha_jG^{2D}_{j}(x)),
\end{equation}
where the $N$ Gaussians overlapping with the pixel $p$ are ordered from front to back, $\mathbf{c_i}$ is defined as the color of each Gaussian, represented as Spherical Harmonics coefficients, and $\alpha_i$ is a learned per-Gaussian scalar opacity.

To optimize all the parameters of each Gaussian, the authors use standard gradient descent with a loss $L$ defined as a combination of the $L_1$ color loss and a D-SSIM term, scaled with a hyperparameter $\lambda$:
\begin{equation}
    L =  (1 - \lambda)L_1 + \lambda L_\text{D-SSIM}.
\end{equation}

To better reconstruct the scene, 3DGS performs an adaptive density control step, cloning Gaussians in the under-reconstructed areas and splitting Gaussians in the over-reconstructed areas. Additionally, to account for potential floaters near the camera and to help control the increase in Gaussians, the opacity values are set to a value close to zero every few thousand iterations.

\subsection{The Symmetric GGX (SGGX) distribution}

In the context of physically based rendering of volumes, it is important to be able to specify angularly varying properties of the medium present in the scene. \Citet{sggx} 
introduced an SSGX distribution that effectively represents the spatially varying properties of anisotropic microflake participating media through a symmetric 3x3 matrix.

In the Canonical Basis, the SGGX distribution can be defined by a $3\times3$ symmetric positive definite matrix of six coefficients:
\begin{equation}
    S = \begin{pmatrix}
        S_{xx}       & S_{xy} & S_{xz} \\
        S_{xy}       & S_{yy} & S_{yz} \\
        S_{xz}       & S_{yz} & S_{zz}
    \end{pmatrix},
    \label{eq:S_SGGX}
\end{equation}

However, the SGGX distribution can also be defined based on the projected area of the microflake in the eigenspace:
\begin{equation}
    S = (\omega_1,\omega_2,\omega_3)\begin{pmatrix}
        S_{11} & 0 & 0 \\
        0      & S_{22} & 0\\
        0      & 0 & S_{33}
    \end{pmatrix}(\omega_1,\omega_2,\omega_3)^T,
\end{equation}
where $S_{11}=\sigma^2(\omega_1)$, $S_{22}=\sigma^2(\omega_2)$ and $S_{33}=\sigma^2(\omega_3)$ are positive eigenvalues that are equal to the squared projected areas of the ellipsoid in the directions given by the orthonormal eigenvectors. Given a matrix $S$ and a direction $\omega_i$, the projected area of the ellipsoid can be computed as:
\begin{equation}
    \sigma(\omega_i) = \int_{\Omega} \langle \omega_i , \omega_m \rangle D(\omega_m) d \omega_m = \sqrt{\omega_i^TS\omega_i},
    \label{eq:sigma_SGGX}
\end{equation}
where $D(\omega_m)$ is defined as the distribution of normals:
\begin{equation}
    D(\omega_m)= \frac{1}{\pi\sqrt{|S|}(\omega_m^TS^{-1}\omega_m)^2}.
\end{equation}

\section{Method}
In our paper, we propose combining the expressive capabilities of the microflake SGGX distribution with 3D Gaussian Splatting to develop a view-dependent opacity model that more accurately captures specular highlights and reflections. We will first define the new opacity model, extended by the symmetric $S$ matrix from the SGGX distribution. Next, we will describe the new density control mechanism, which regulates the number of Gaussians through pruning and densification techniques. Then, we will propose an updated opacity rest system, responsible for minimizing the number of floaters, accounting for our extended model. Finally, we will propose a regularization loss that enforces better view consistency of our 3D Gaussians. Please see Figure~\ref{fig:teaser} for a high overview of the method.

\subsection{View-dependent opacity through a symmetric matrix}
In the standard 3DGS implementation, the opacity of a 3D Gaussian $i$ is defined by a learnable per-Gaussian scalar $\gamma_i$, where the final opacity $\alpha_i$ can be represented as:
\begin{equation}
    \alpha_i = \sigma(\gamma_i),
\end{equation}
where $\sigma$ here denotes the sigmoid function. We propose to extend the definition of the scalar opacity $\alpha_i$ by introducing a symmetric matrix $\hat{S}_i$, parametrized with six learnable components, like the $S$ matrix in the Canonical Basis in Eq.~\ref{eq:S_SGGX}. To better capture the view-dependent light interactions, like specular highlights and reflections, without compromising the appearance of diffuse surfaces, we propose to combine both the parameter $\gamma_i$ with the newly introduced matrix $\hat{S}_i$. The opacity function $\hat{\alpha}_i$ can be defined as:
\begin{equation}
    \hat{\alpha_i}(\omega_{i,j}) = \sigma(\gamma_{i,j} + \omega_{i,j}^T \hat{S}_i \omega_{i,j}),
\end{equation}
where $\omega_{i,j}$ defines the view vector between the center of the Gaussian $i$ and the center of the camera $j$, allowing to define a symmetric angular distribution, similar to the SSGX distribution \cite{sggx}.

\subsection{Density control in the view-dependent setting}
In the original 3DGS paper, during the density control step of the optimization, the Gaussians were actively pruned when their opacity was lower than an introduced hyperparameter $\tau$: $\alpha < \tau$.

To allow for opacity-based pruning within our method as well, we propose to prune the Gaussians whose maximum opacity in the training set $N$ is lower than $\tau$:
\begin{equation}
    \hat{\alpha}(\omega_{i,j^*}) < \tau,
\end{equation}
where $j^* = \arg\max_{j\in N} \hat{\alpha_i}(\omega_{i,j})$.

\subsection{Enhanced opacity reset}
In the original implementation of the paper, the opacity of the 3D Gaussians is reset to a value close to zero every few thousand iterations to minimize the number of Gaussians and lessen the impact of possible floaters. To facilitate a similar approach, we propose resetting the per-Gaussian matrix $\hat{S}_i$ by computing its eigendecomposition and calculating a new matrix $\hat{S}^*_i$ as either
\begin{align}
    \label{eq:vod-3dgs}
    \hat{S}_i &= Q \begin{pmatrix}
        \lambda_0 & 0 & 0 \\
        0 & 0 & 0 \\
        0 & 0 & 0
    \end{pmatrix} Q^{-1},
    &&\text{or}&&
    \hat{S}_i = Q \begin{pmatrix}
        0 & 0 & 0 \\
        0 & 0 & 0 \\
        0 & 0 & \lambda_2
    \end{pmatrix} Q^{-1},
\end{align}
where $\lambda_0$ and $\lambda_2$ are the highest and lowest eigenvalues respectively, and $Q$ is a matrix of eigenvectors. We refer to these as \textbf{VoD-3DGS[L]} and \textbf{VoD-3DGS[S]} respectively in the Section~\ref{sec:results}. We decided to include both opacity resetting methods as they produce viable results in various datasets.

\subsection{View consistency regularization loss}
Finally, we propose to extend the loss function used for optimization in the standard 3DGS setting. We introduce a new view consistency loss $L_{vc}$ that allows the same Gaussian, viewed from similar angles, to maintain a similar opacity. 

Given a camera view $i$, we stochastically sample another view $j$, where the likelihood of selecting a specific view $j$ is based on the number of keypoint matches between both views $i$ and $j$. Then, we compute our loss $L_{vc}$ for all the Gaussians $g \in G$ as:
\begin{equation}
    L_{vc} = \frac{1}{|G|}\sum_{g \in G}\left[\max(\cos(\theta),0)(\hat{\alpha}(\omega_{g,i})-\hat{\alpha}(\omega_{g,j}))^2\right],
\end{equation}
where $\theta$ is defined as the angle between the view vector $\omega_{g,i}$ and $\omega_{g,j}$.

Therefore, the final optimization loss can be represented as:
\begin{equation}
    L =  (1 - \lambda)L_1 + \lambda L_{D-SSIM} +  L_{vc}.
\end{equation}

\section{Implementation, Evaluation and Results}
\label{sec:results}

\subsection{Implementation}
We implement our VoD-3DGS upon the original codebase of the original 3D Gaussian Splatting paper. 

We do not modify any existing parameters, except for the learning rate of the $\gamma$ parameter responsible for opacity learning. To learn the per-Gaussian symmetric matrix $\hat{S_i}$, we set the learning rate for each of the six components and the $\gamma$ term to $\frac{1}{4} lr_{\gamma}$, where $lr_\gamma$ is the original learning rate used in the 3D Gaussian Splatting paper to optimize $\gamma$. 

As a preprocessing step, to create a per-view distribution for sampling views used in our view-consistency loss $L_{vc}$, we use LightGlue \cite{lightglue} to compute keypoint matches between pairs of images. However, since standard 3D Gaussian splatting already utilizes Structure-from-Motion (SfM) \cite{sfm}, which also computes the information about keypoint matches between images, all the necessary information could be collected during the SfM computation.

We conduct all the experiments on a single 48GB NVIDIA L40 GPU.

\subsection{Datasets}
We evaluate the performance of our proposed algorithm on four datasets: 
\begin{enumerate}
    \item MipNerf-360 \cite{mipnerf360},
    \item Tanks \& Temples \cite{tanksAndtemples},
    \item Deep Blending  \cite{deepBlending},
    \item and NeRF-Synthetic \cite{nerf}. 
\end{enumerate}

\begin{table*}[ht]
\caption{A quantitative comparison of our method was computed over three datasets compared to previous work. Our method achieves higher-quality results for all datasets in almost all the metrics for novel view synthesis while achieving $>60$ FPS rendering performance with reasonable memory utilization. Results marked with \textdagger have been recomputed and \textdaggerdbl have been adapted from \cite{3dcs}, where other values have been directly adopted from the original papers. As for the training times, we report the training time with and without the keypoint pre-computation step, indicated by the number outside and in the brackets respectively.}
\label{tab:results_real}
\centering
\resizebox{\textwidth}{!}{%
\begin{tabular}{c|cccccccccccccccccc|}
\cline{2-19}
 &
  \multicolumn{18}{c|}{\textbf{Dataset}} \\ \hline
\multicolumn{1}{|c|}{} &
  \multicolumn{6}{c|}{Mip-NeRF360 Dataset} &
  \multicolumn{6}{c|}{Tanks \& Temples} &
  \multicolumn{6}{c|}{Deep Blending} \\
\multicolumn{1}{|c|}{\multirow{-2}{*}{\textbf{Method}}} &
  \textit{LPIPS $\downarrow$} &
  \textit{PSNR $\uparrow$} &
  \textit{SSIM $\uparrow$} &
  Train &
  FPS &
  \multicolumn{1}{c|}{Mem} &
  \textit{LPIPS $\downarrow$} &
  \textit{PSNR $\uparrow$} &
  \textit{SSIM $\uparrow$} &
  Train &
  FPS &
  \multicolumn{1}{c|}{Mem} &
  \textit{LPIPS $\downarrow$} &
  \textit{PSNR $\uparrow$} &
  \textit{SSIM $\uparrow$} &
  Train &
  FPS &
  Mem \\ \midrule
\multicolumn{1}{|c|}{Mip-Nerf360} &
  0.237 &
  \cellcolor[HTML]{F8CD99}27.69 &
  0.792 &
  48h &
  0.06 &
  \multicolumn{1}{c|}{8.6MB} &
  0.257 &
  22.22 &
  0.759 &
  48h &
  0.14 &
  \multicolumn{1}{c|}{8.6MB} &
  0.245 &
  29.40 &
  0.901 &
  48h &
  0.09 &
  8.6MB \\
\multicolumn{1}{|c|}{GES} &
  0.250 &
  26.91 &
  0.794 &
  32m &
  186 &
  \multicolumn{1}{c|}{377MB} &
  0.198 &
  23.35 &
  0.836 &
  21m &
  210 &
  \multicolumn{1}{c|}{222MB} &
  0.252 &
  29.68 &
  0.901 &
  30m &
  160 &
  399MB \\
\multicolumn{1}{|c|}{3DGS} &
  \cellcolor[HTML]{FCF9AD}0.214 &
  27.21 &
  0.815 &
  42m &
  134 &
  \multicolumn{1}{c|}{734MB} &
  0.183 &
  23.14 &
  0.841 &
  26m &
  154 &
  \multicolumn{1}{c|}{411MB} &
  0.243 &
  29.41 &
  \cellcolor[HTML]{FCF9AD}0.903 &
  36m &
  137 &
  676MB \\
\multicolumn{1}{|c|}{2DGS  \textdaggerdbl} &
  0.252 &
  27.18 &
  0.808 &
  29m &
  64 &
  \multicolumn{1}{c|}{484MB} &
  0.212 &
  23.13 &
  0.831 &
  14m &
  122 &
  \multicolumn{1}{c|}{200MB} &
  0.257 &
  29.50 &
  0.902 &
  28m &
  76 &
  353MB \\
\multicolumn{1}{|c|}{3DCS} &
  \cellcolor[HTML]{F5999A}0.207 &
  27.29 &
  0.802 &
  87m &
  25 &
  \multicolumn{1}{c|}{666MB} &
  \cellcolor[HTML]{F5999A}0.157 &
  \cellcolor[HTML]{F8CD99}23.95 &
  0.851 &
  60m &
  33 &
  \multicolumn{1}{c|}{282MB} &
  \cellcolor[HTML]{F5999A}0.237 &
  \cellcolor[HTML]{F8CD99}29.81 &
  0.902 &
  71m &
  30 &
  332MB \\
\multicolumn{1}{|c|}{3DGS \textdagger} &
  0.215 &
  27.53 &
  \cellcolor[HTML]{FCF9AD}0.816 &
  31m (32m) &
  113 &
  \multicolumn{1}{c|}{649MB} &
  0.169 &
  23.75 &
  \cellcolor[HTML]{FCF9AD}0.852 &
  19m (20m) &
  155 &
  \multicolumn{1}{c|}{372MB} &
  \cellcolor[HTML]{F8CD99}{\color[HTML]{333333} 0.238} &
  29.77 &
  \cellcolor[HTML]{F8CD99}0.907 &
  31m (31m) &
  117 &
  586MB \\ \hline
\multicolumn{1}{|c|}{\textbf{VoD-3DGS{[}S{]}}} &
  0.217 &
  \cellcolor[HTML]{FCF9AD}27.66 &
  \cellcolor[HTML]{F8CD99}0.817 &
  47m (73m) &
  78 &
  \multicolumn{1}{c|}{687MB} &
  \cellcolor[HTML]{FCF9AD}0.165 &
  \cellcolor[HTML]{F5999A}{\color[HTML]{333333} 24.17} &
  \cellcolor[HTML]{F8CD99}{\color[HTML]{333333} 0.859} &
  28m (65m) &
  106 &
  \multicolumn{1}{c|}{430MB} &
  0.244 &
  \cellcolor[HTML]{FCF9AD}29.75 &
  \cellcolor[HTML]{F8CD99}0.907 &
  38m (70m) &
  91 &
  559MB \\
\multicolumn{1}{|c|}{\textbf{VoD-3DGS{[}L{]}}} &
  \cellcolor[HTML]{F8CD99}0.213 &
  \cellcolor[HTML]{F5999A}{\color[HTML]{333333} 27.79} &
  \cellcolor[HTML]{F5999A}0.818 &
  53m (83m) &
  65 &
  \multicolumn{1}{c|}{849MB} &
  \cellcolor[HTML]{F8CD99}0.160 &
  \cellcolor[HTML]{FCF9AD}{\color[HTML]{333333} 23.91} &
  \cellcolor[HTML]{F5999A}0.860 &
  31m (71m) &
  84 &
  \multicolumn{1}{c|}{525MB} &
  \cellcolor[HTML]{FCF9AD}0.240 &
  \cellcolor[HTML]{F5999A}29.84 &
  \cellcolor[HTML]{F5999A}{\color[HTML]{333333} 0.908} &
  39m (70m) &
  85 &
  606MB \\ \bottomrule
\end{tabular}%
}
\end{table*}

\begin{table}[ht]
\caption{A quantitative evaluation of our method compared to previous work was computed over the Nerf-Synthethic \cite{nerf} dataset. Results marked with \textdagger have been recomputed, where other values have been directly adopted from the original papers.}
\begin{tabular}{c|ccc|}
\cline{2-4}
                                                      & \multicolumn{3}{c|}{\textbf{Dataset - Nerf Synthethic}}                                        \\ \hline
\multicolumn{1}{|c|}{Method}                          & \textit{LPIPS $\downarrow$}    & \textit{PSNR $\uparrow$}      & \textit{SSIM $\uparrow$}      \\ \midrule
\multicolumn{1}{|c|}{Mip-Nerf 360}                    & 0.060                          & 30.34                         & 0.951                         \\
\multicolumn{1}{|c|}{Zip-Nerf}                        & 0.031                          & 33.10                         & \cellcolor[HTML]{F5999A}0.971 \\ \hline
\multicolumn{1}{|c|}{3DGS}                            & 0.037  & 33.31 & \cellcolor[HTML]{FCF9AD}0.969 \\
\multicolumn{1}{|c|}{GOF}                             & 0.038                          & 33.45                         & 0.969                         \\
\multicolumn{1}{|c|}{GS Shader}                       & \cellcolor[HTML]{F5999A}0.029  & 33.38 & 0.968 \\
\multicolumn{1}{|c|}{3DGS \textdagger} & 0.0298                         & \cellcolor[HTML]{FCF9AD}33.48 & \cellcolor[HTML]{F8CD99}0.970 \\ \hline
\multicolumn{1}{|c|}{\textbf{VoD-3DGS{[}S{]}}}          & \cellcolor[HTML]{F8CD99}0.0296 & \cellcolor[HTML]{F5999A}33.79 & \cellcolor[HTML]{F5999A}0.971 \\
\multicolumn{1}{|c|}{\textbf{VoD-3DGS{[}L{]}}}          & \cellcolor[HTML]{FCF9AD}0.0297 & \cellcolor[HTML]{F8CD99}33.69 & \cellcolor[HTML]{F5999A}0.971 \\ \bottomrule
\end{tabular}%
\label{tab:results_synth}
\end{table}

\subsection{Evaluation}
We use the same training process for evaluation as 3DGS \cite{3DGS}, using all the scenes in the MipNerf360 dataset \cite{mipnerf360}, the Truck and Train scene from the Tanks \& Temples dataset \cite{tanksAndtemples}, and the Playroom and Dr. Johnson scenes from the Deep Blending dataset \cite{deepBlending}. Additionally, we provide results on all the scenes from the NeRF-Synthetic dataset \cite{nerf} to showcase our algorithm's performance in the synthetic scenarios. We compute PSNR, SSIM, and LPIPS for each scene on the unseen testing dataset and report the mean over all the scenes used per dataset. Additionally, we report per-dataset mean FPS count, memory used, and training time for the real-life datasets.

We present our results under two names: \textbf{VoD-3DGS[S]} and \textbf{VoD-3DGS[L]}. The only distinction being whether we use the smallest or largest eigenvalue to reset the $S$ matrix of each Gaussian (Eq.~\ref{eq:vod-3dgs}). The two names represent the relative memory used, where VoD-3DGS[L] has a higher memory footprint than VoD-3DGS[S].

We present our findings for the three real-life datasets in Table\ref{tab:results_real} and the synthetic dataset in Table~\ref{tab:results_synth}. As our baselines, we use the state-of-the-art NeRF approaches (Mip-Nerf360 \cite{mipnerf360} and Zip-Nerf \cite{zipnerf}), 3D Gaussian Splatting \cite{3DGS}, and 3D Gaussian Splatting methods that enhance the 3D Gaussian native model without any neural network introduced for the representation where the results are provided on the original parameters of the 3DGS (2DGS \cite{2dgs}, GES \cite{ges}, GOF \cite{gof}, 3DCS \cite{3dcs} and GS Shader \cite{gs-shader}). Moreover, we provid a separate evaluation of our approach on Spec-Gaussian \cite{spec-gaussian} in Section~\ref{sec:spec}.

\subsection{Results}
\begin{figure*}[ht]
    \centering
    \begin{subfigure}{\textwidth}
        \begin{subfigure}{0.33\linewidth}
            \centering
            \begin{subfigure}{\linewidth}
                \centering
                \includegraphics[width=1\linewidth]{gt_garden_00010_boxed.png}
            \end{subfigure}
            \begin{subfigure}{\linewidth}
                \centering
                \includegraphics[width=1\linewidth]{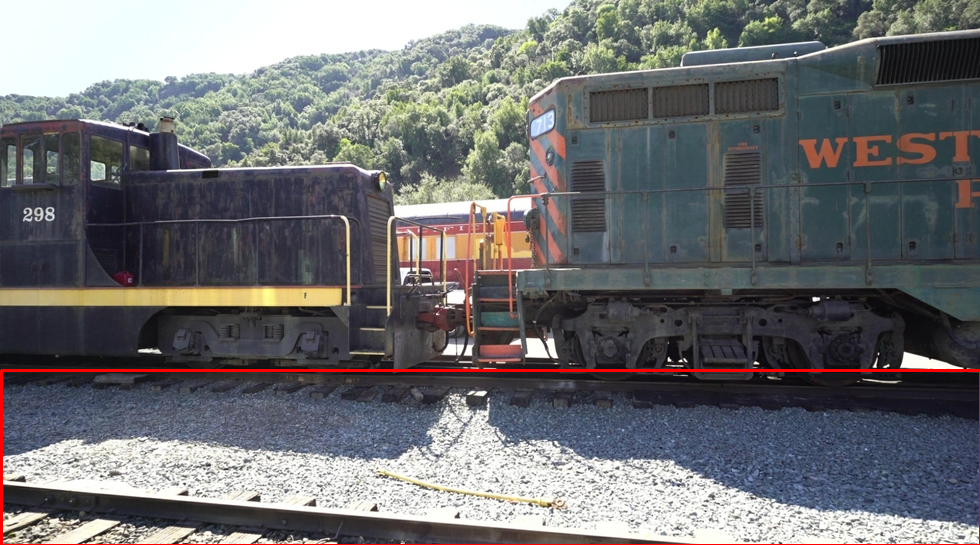}
            \end{subfigure}
            \begin{subfigure}{\linewidth}
                \centering
                \includegraphics[width=1\linewidth]{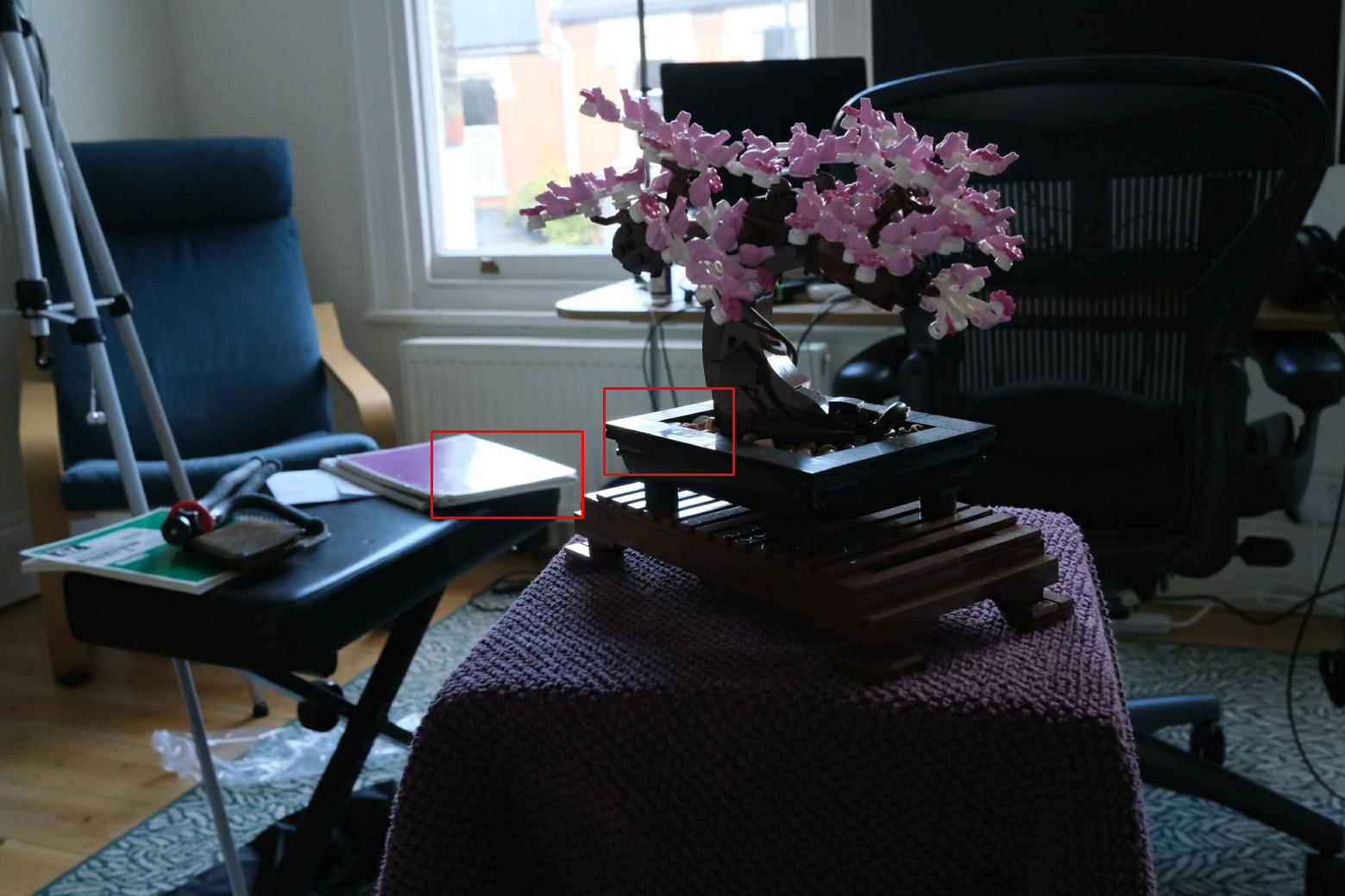}
            \end{subfigure}
            \begin{subfigure}{\linewidth}
                \centering
                \includegraphics[width=1\linewidth]{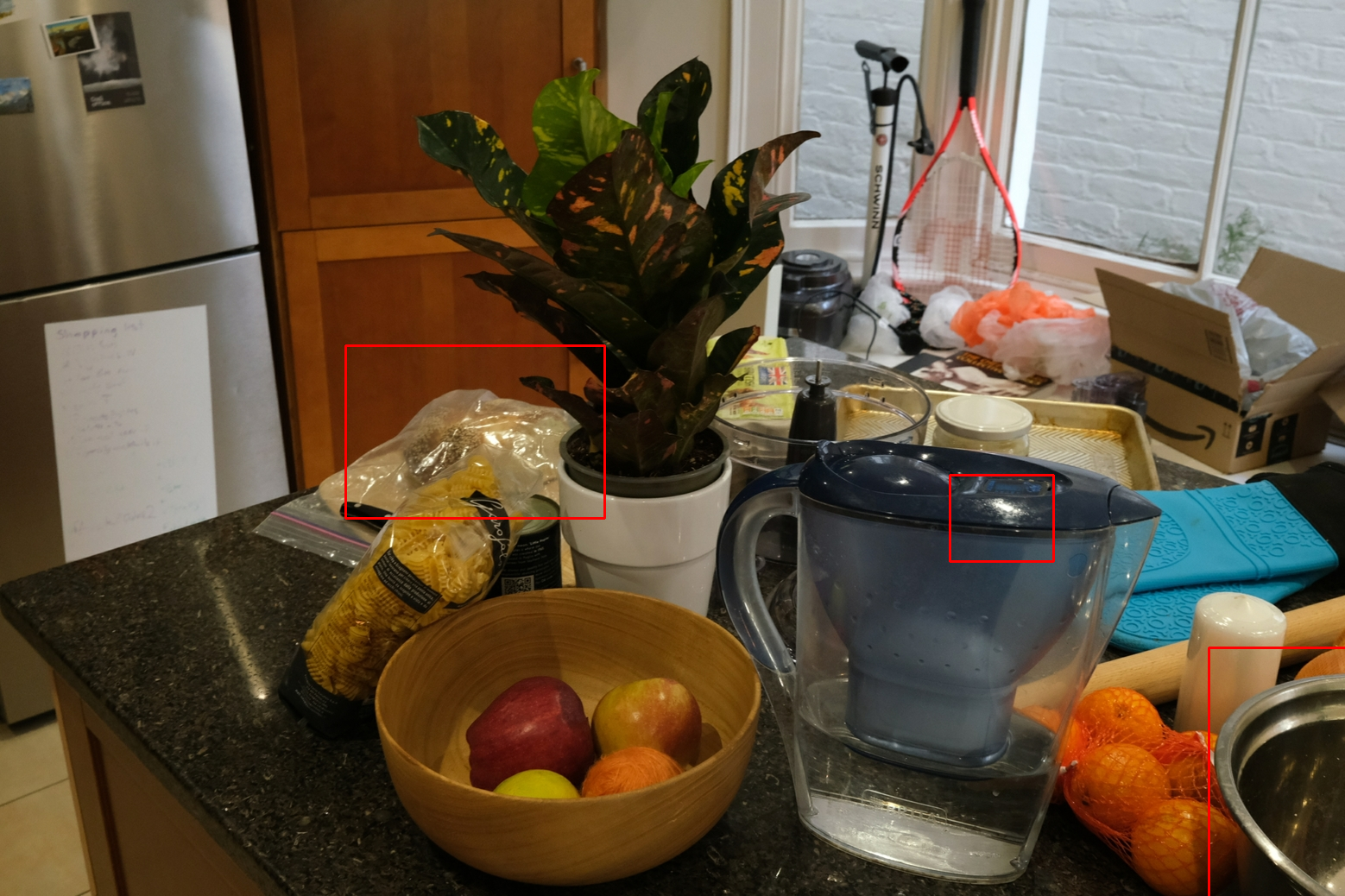}
            \end{subfigure}
            \begin{subfigure}{\linewidth}
                \centering
                \includegraphics[width=1\linewidth]{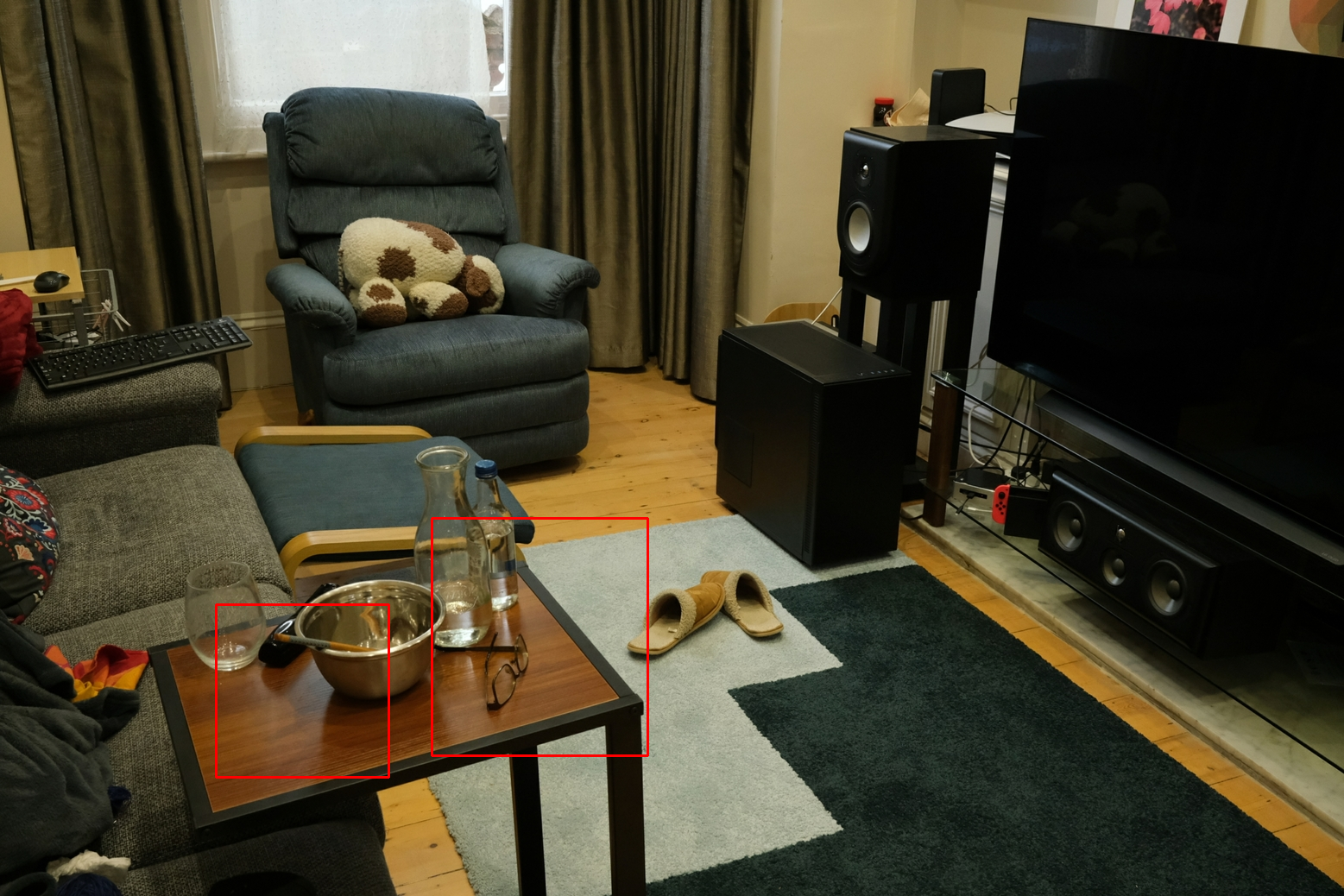}
            \end{subfigure}
            \subcaption{Ground Truth}   
        \end{subfigure}
        \begin{subfigure}{0.33\linewidth}
            \centering
            \begin{subfigure}{\linewidth}
                \centering
                \includegraphics[width=1\linewidth]{3DGS_garden_00010_boxed.png}
            \end{subfigure}
            \begin{subfigure}{\linewidth}
                \centering
                \includegraphics[width=1\linewidth]{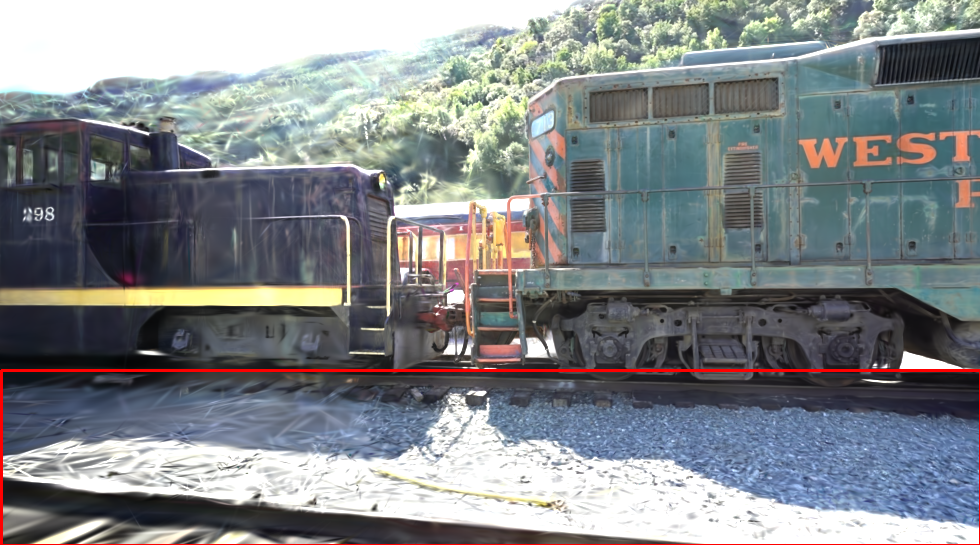}
            \end{subfigure}
            \begin{subfigure}{\linewidth}
                \centering
                \includegraphics[width=1\linewidth]{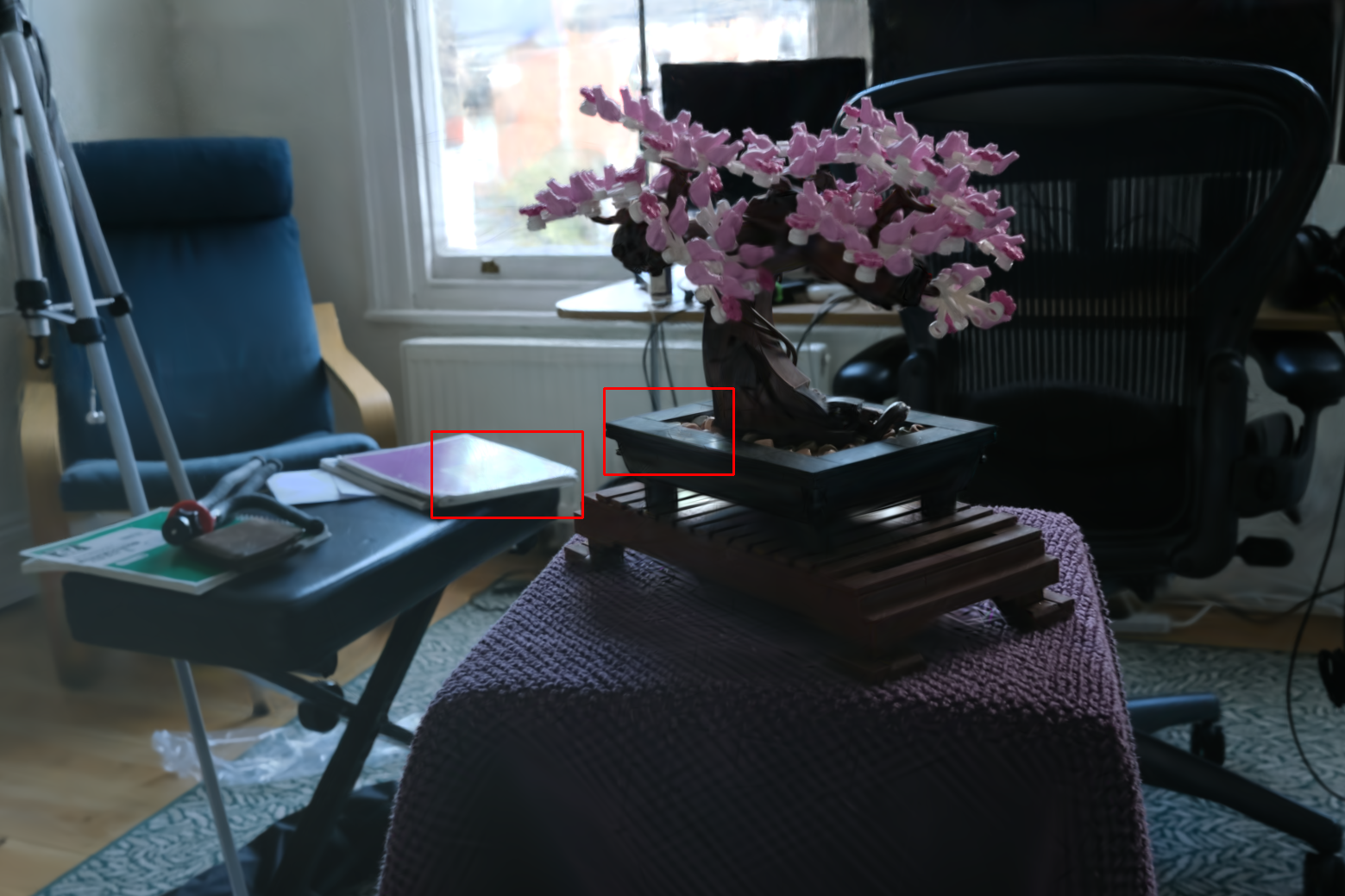}
            \end{subfigure}
            \begin{subfigure}{\linewidth}
                \centering
                \includegraphics[width=1\linewidth]{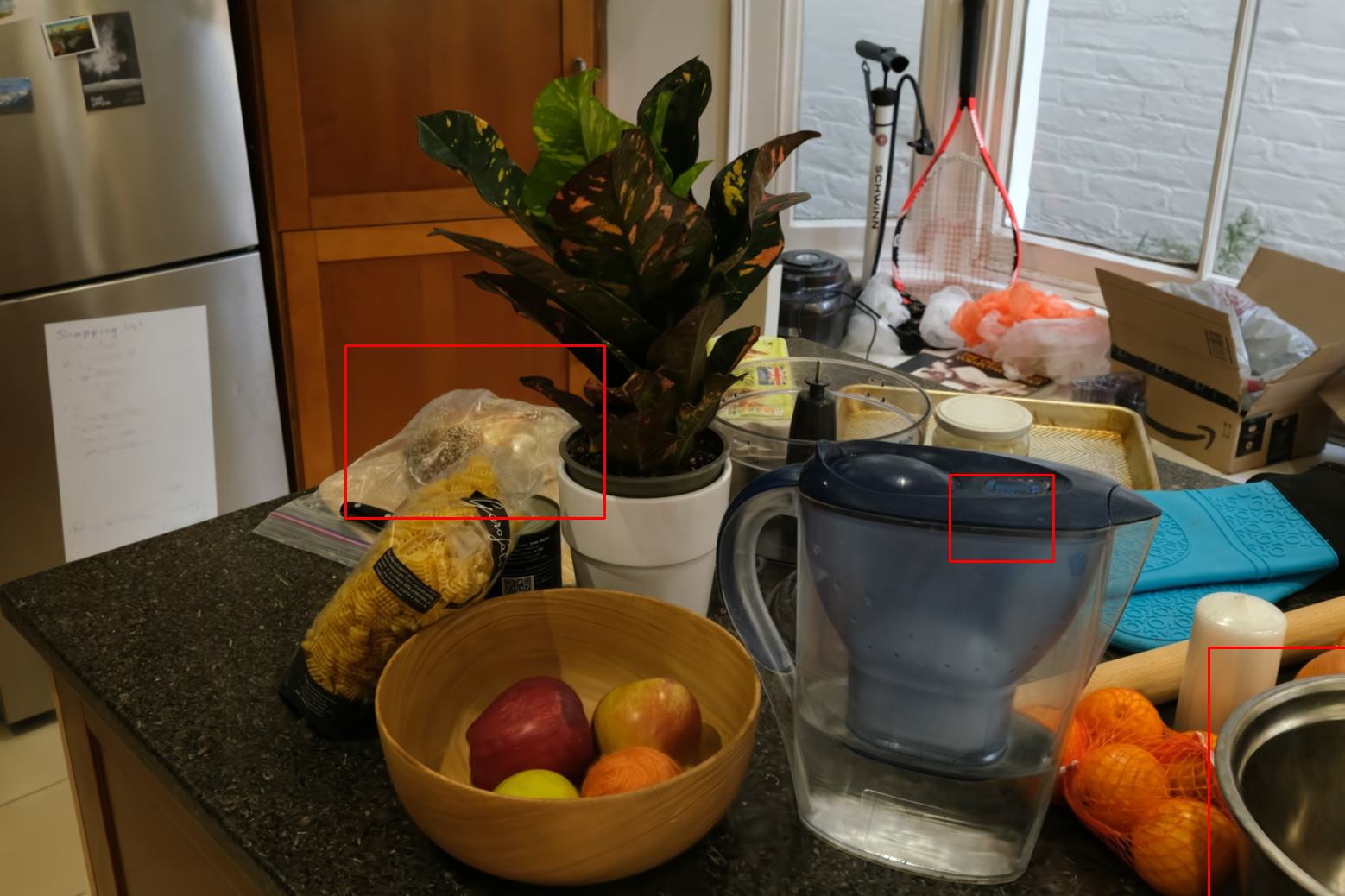}
            \end{subfigure}
            \begin{subfigure}{\linewidth}
                \centering
                \includegraphics[width=1\linewidth]{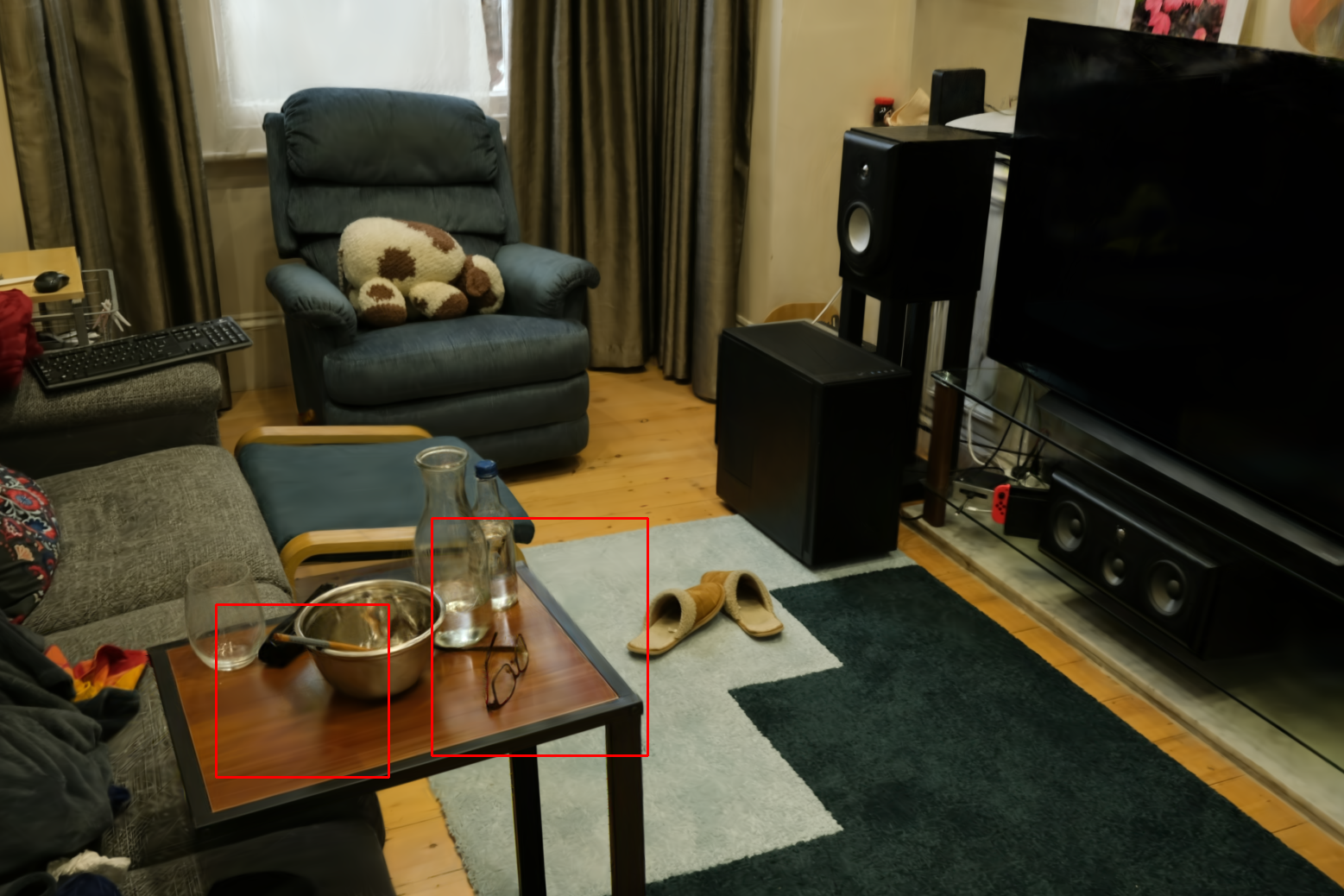}
            \end{subfigure}
            \subcaption{3DGS}   
        \end{subfigure}
        \begin{subfigure}{0.33\linewidth}
            \centering
            \begin{subfigure}{\linewidth}
                \centering
                \includegraphics[width=1\linewidth]{us_L_garden_00010_boxed.png}
            \end{subfigure}
            \begin{subfigure}{\linewidth}
                \centering
                \includegraphics[width=1\linewidth]{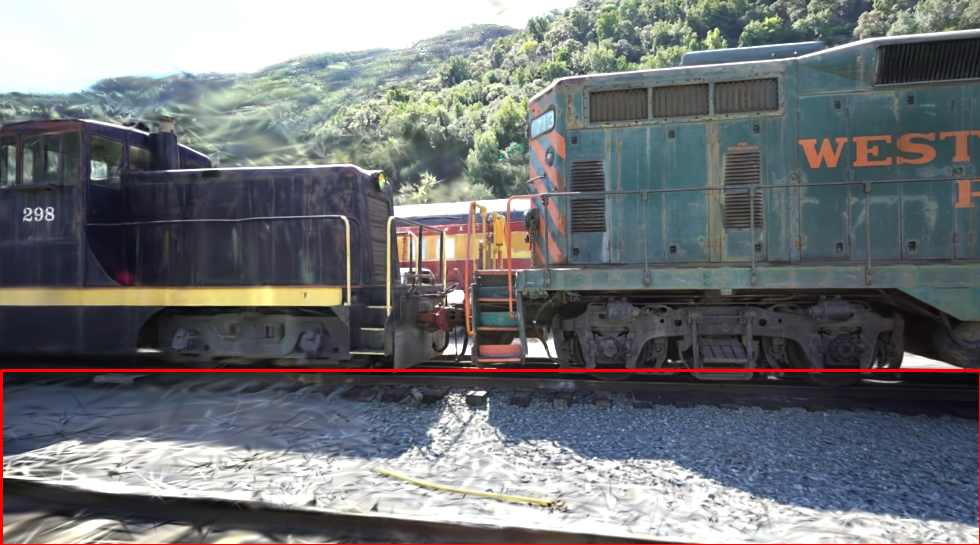}
            \end{subfigure}
            \begin{subfigure}{\linewidth}
                \centering
                \includegraphics[width=1\linewidth]{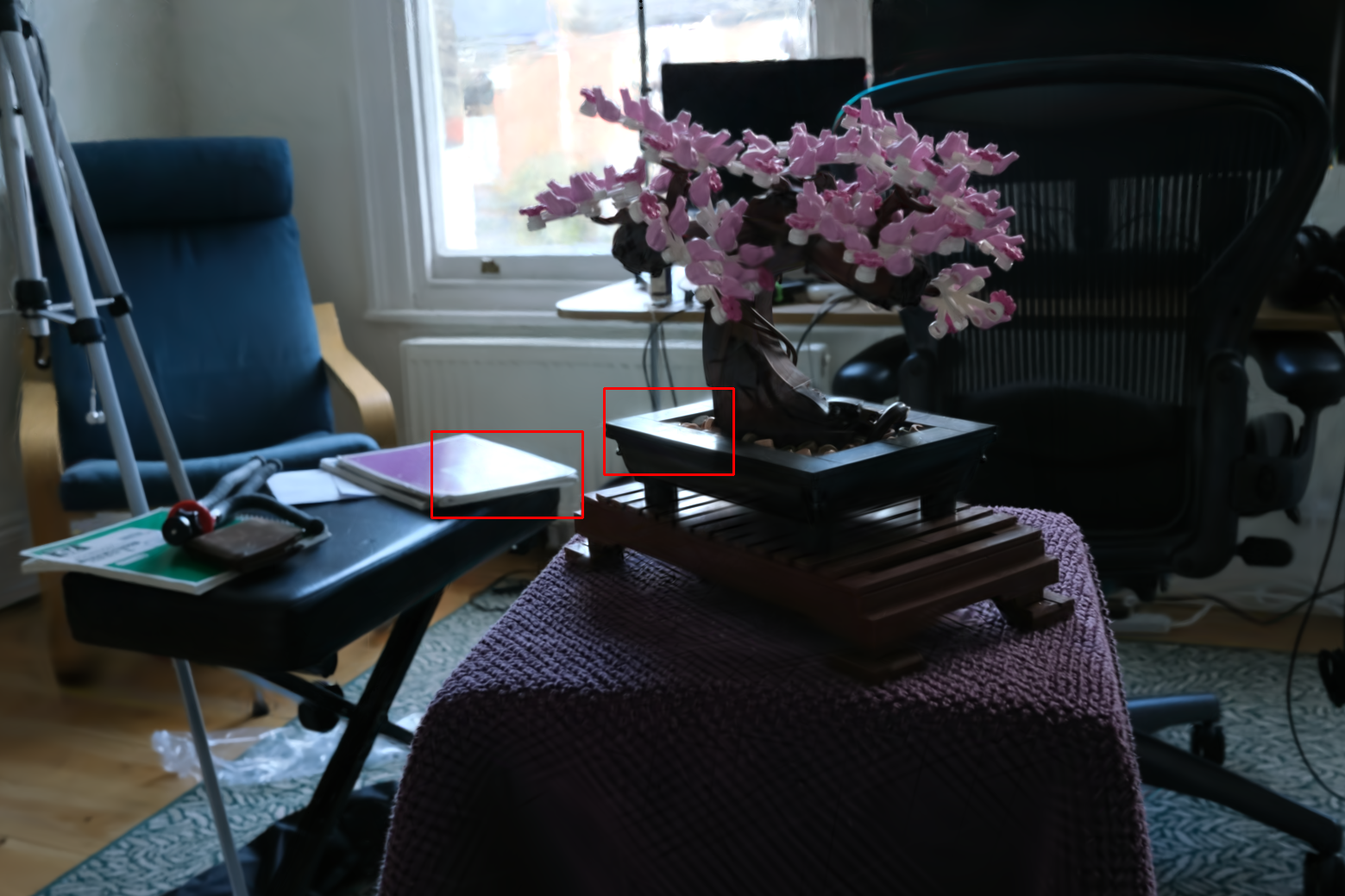}
            \end{subfigure}
            \begin{subfigure}{\linewidth}
                \centering
                \includegraphics[width=1\linewidth]{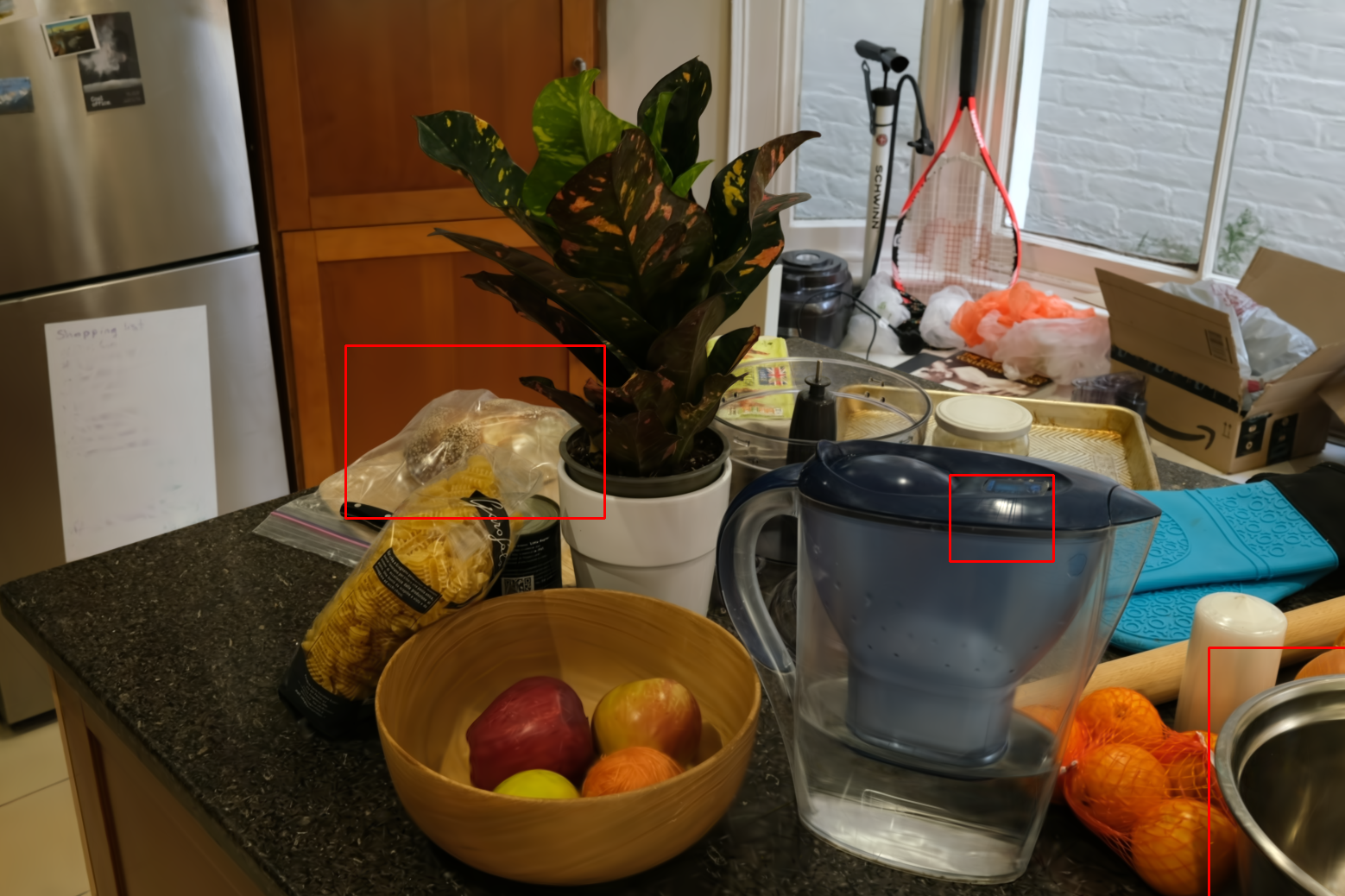}
            \end{subfigure}
            \begin{subfigure}{\linewidth}
                \centering
                \includegraphics[width=1\linewidth]{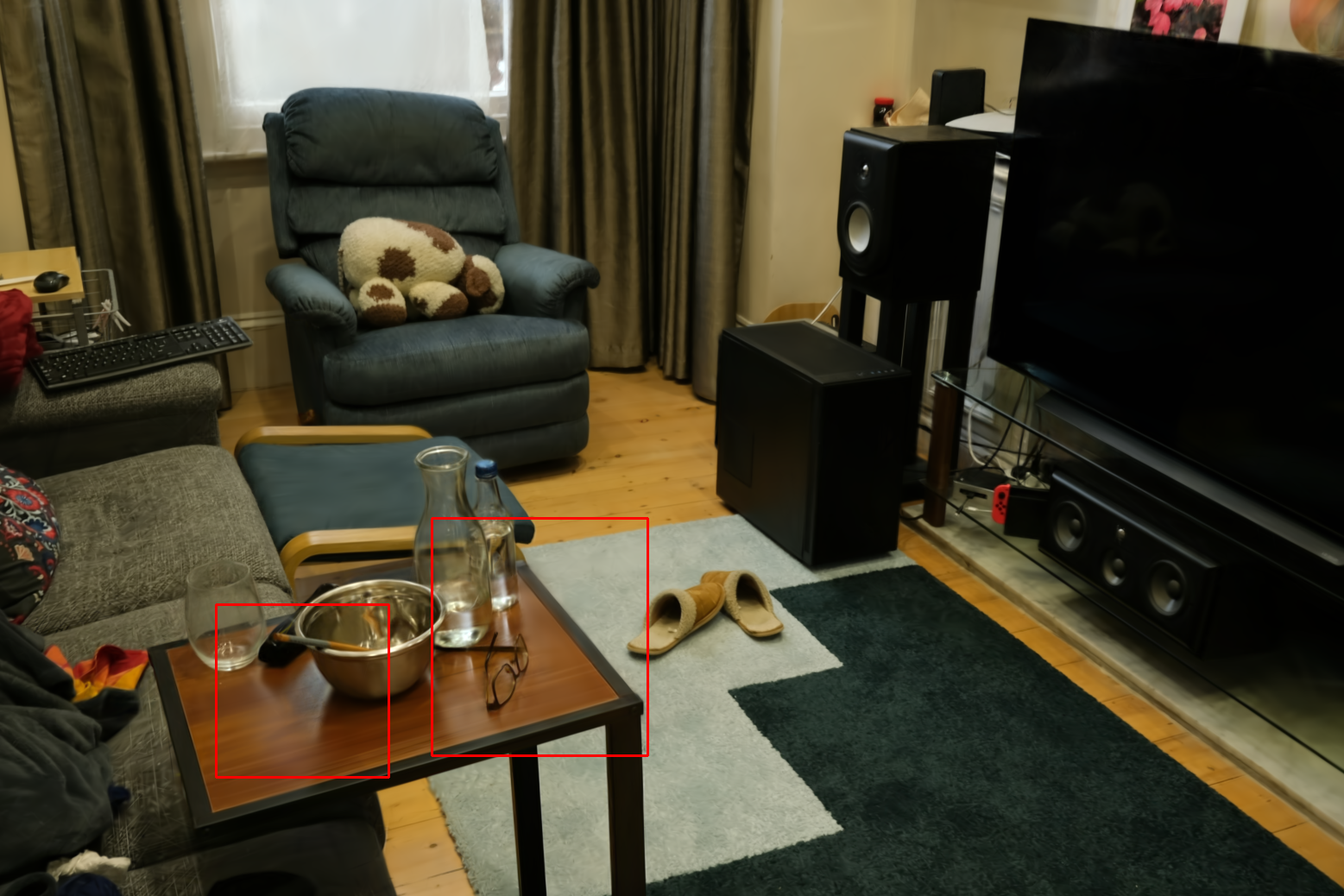}
            \end{subfigure}
            \subcaption{VoD-3DGS[L] (Our method)}   
        \end{subfigure}
    \end{subfigure} 
  \caption{Comparison of our proposed method to standard 3DGS. Our method boosts the specular response in various scenes, which can be seen on the table
center (Garden scene; first row), the water filter (Counter scene; fourth row), the rightmost corner of the table (Room scene; last row), and the left side of the base
of the bonsai (Bonsai scene; third row). Our method also allows changing light conditions (white gravel and train tracks in the Train scene; second
row).}
  \label{fig:full_page_one}
\end{figure*}

\begin{figure*}[ht]
    \centering
    \begin{subfigure}{\textwidth}
        \begin{subfigure}{0.33\linewidth}
            \centering
            \begin{subfigure}{\linewidth}
                \centering
                \includegraphics[width=1\linewidth]{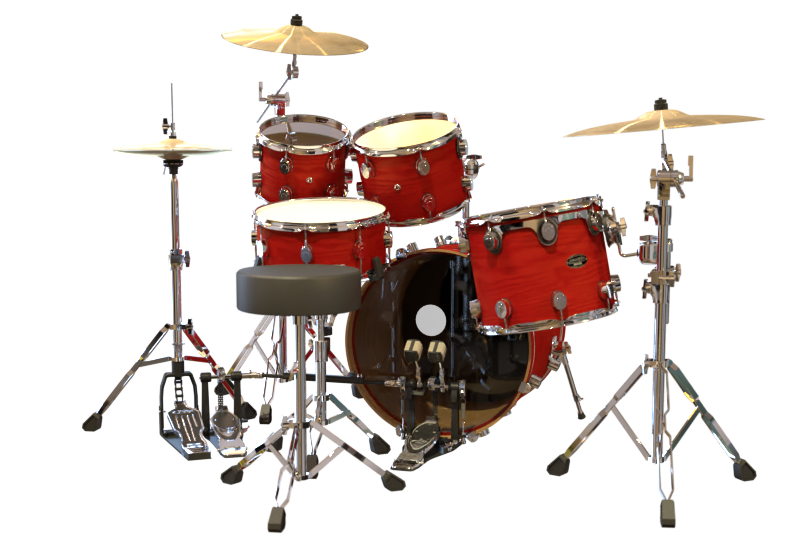}
            \end{subfigure}
            \begin{subfigure}{0.49\linewidth}
                \centering
                \includegraphics[width=\linewidth]{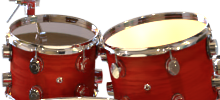}
            \end{subfigure}
            \begin{subfigure}{0.49\linewidth}
                \centering
                \includegraphics[width=\linewidth]{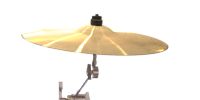}
            \end{subfigure}
            \begin{subfigure}{\linewidth}
                \centering
                \includegraphics[width=0.9\linewidth]{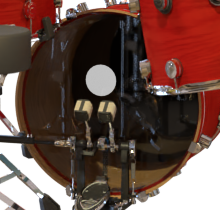}
            \end{subfigure}
            \begin{subfigure}{0.74\linewidth}
                \centering
                \includegraphics[width=\linewidth]{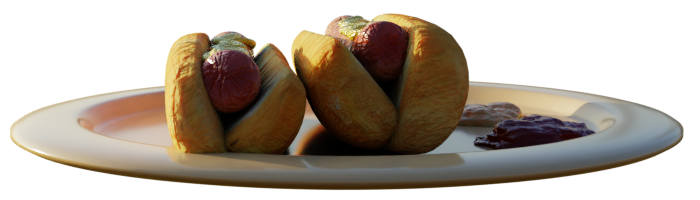}
            \end{subfigure}
            \begin{subfigure}{0.24\linewidth}
                \centering
                \includegraphics[width=\linewidth]{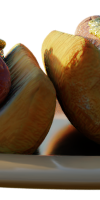}
            \end{subfigure}
            \begin{subfigure}{\linewidth}
                \centering
                \includegraphics[width=1\linewidth]{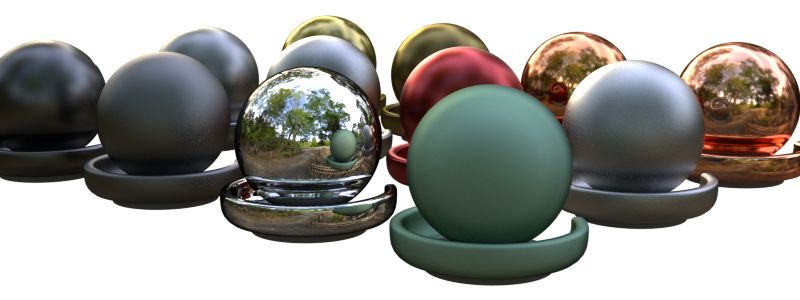}
            \end{subfigure}
            \begin{subfigure}{\linewidth}
                \centering
                \includegraphics[width=1\linewidth]{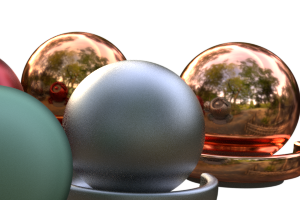}
            \end{subfigure}
            \subcaption{Ground Truth}   
        \end{subfigure}
        \begin{subfigure}{0.33\linewidth}
            \centering
            \begin{subfigure}{\linewidth}
                \centering
                \includegraphics[width=1\linewidth]{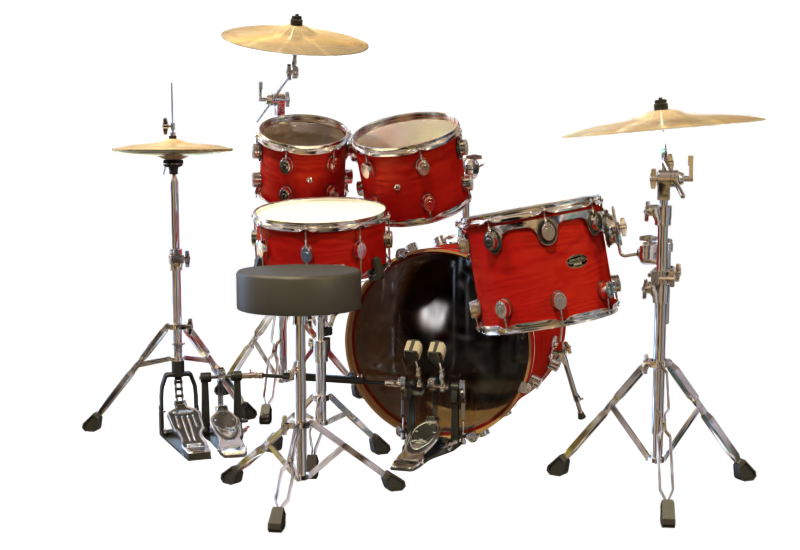}
            \end{subfigure}
            \begin{subfigure}{0.49\linewidth}
                \centering
                \includegraphics[width=\linewidth]{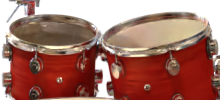}
            \end{subfigure}
            \begin{subfigure}{0.49\linewidth}
                \centering
                \includegraphics[width=\linewidth]{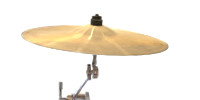}
            \end{subfigure}
            \begin{subfigure}{0.9\linewidth}
                \centering
                \includegraphics[width=\linewidth]{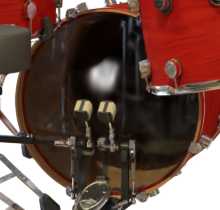}
            \end{subfigure}
            \begin{subfigure}{0.74\linewidth}
                \centering
                \includegraphics[width=\linewidth]{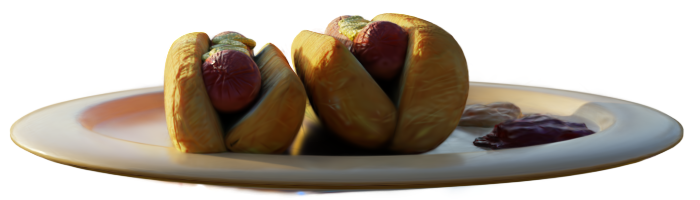}
            \end{subfigure}
            \begin{subfigure}{0.24\linewidth}
                \centering
                \includegraphics[width=\linewidth]{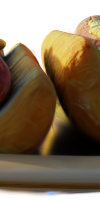}
            \end{subfigure}
            \begin{subfigure}{\linewidth}
                \centering
                \includegraphics[width=1\linewidth]{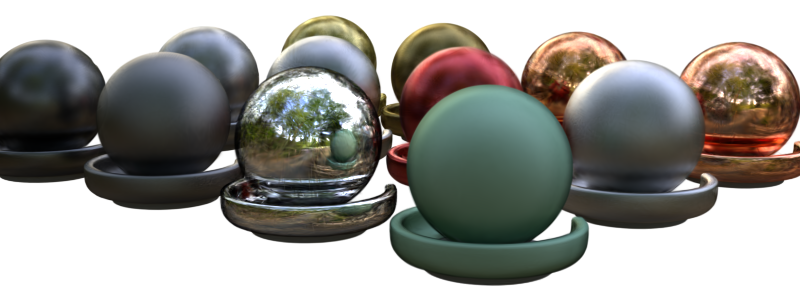}
            \end{subfigure}
                \begin{subfigure}{\linewidth}
                \centering
                \includegraphics[width=1\linewidth]{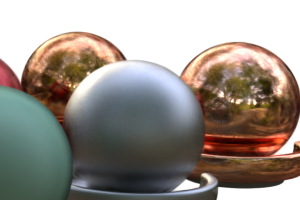}
            \end{subfigure}
            \subcaption{3DGS}   
        \end{subfigure}
        \begin{subfigure}{0.33\linewidth}
            \centering
            \begin{subfigure}{\linewidth}
                \centering
                \includegraphics[width=1\linewidth]{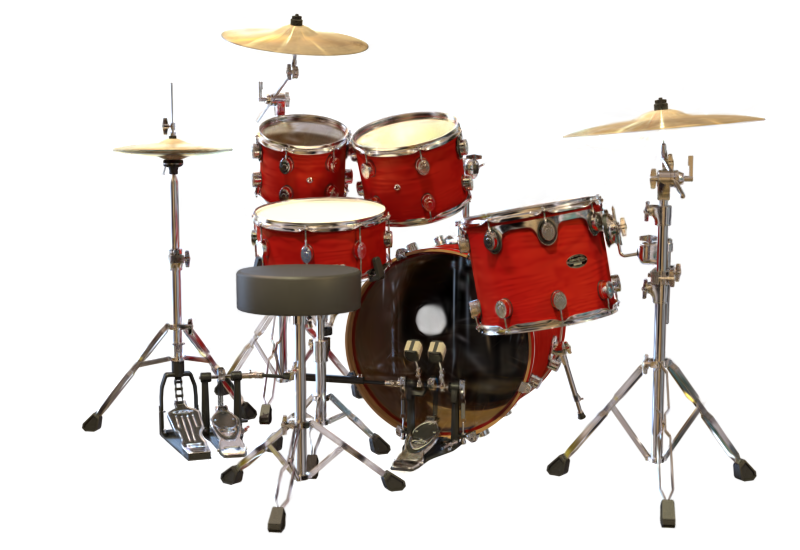}
            \end{subfigure}
            \begin{subfigure}{0.49\linewidth}
                \centering
                \includegraphics[width=\linewidth]{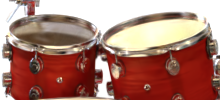}
            \end{subfigure}
            \begin{subfigure}{0.49\linewidth}
                \centering
                \includegraphics[width=\linewidth]{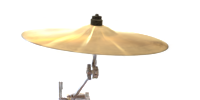}
            \end{subfigure}
            \begin{subfigure}{0.9\linewidth}
                \centering
                \includegraphics[width=\linewidth]{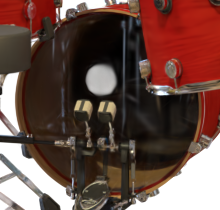}
            \end{subfigure}
            \begin{subfigure}{0.74\linewidth}
                \centering
                \includegraphics[width=\linewidth]{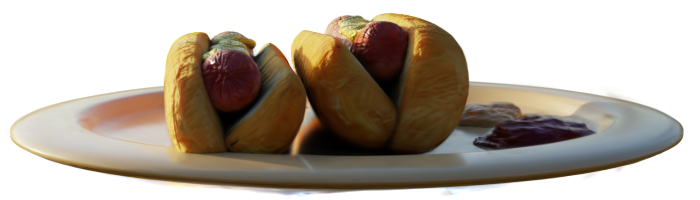}
            \end{subfigure}
            \begin{subfigure}{0.24\linewidth}
                \centering
                \includegraphics[width=\linewidth]{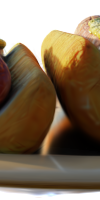}
            \end{subfigure}
            \begin{subfigure}{\linewidth}
                \centering
                \includegraphics[width=1\linewidth]{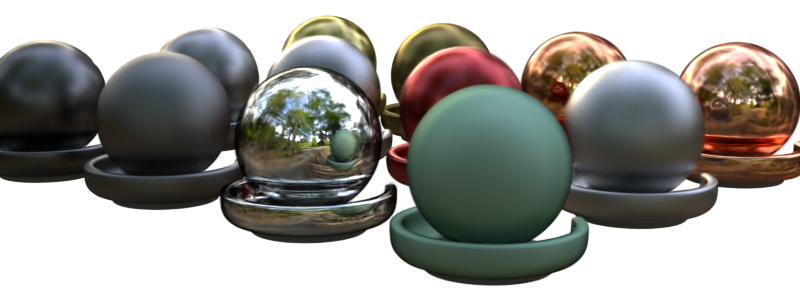}
            \end{subfigure}
            \begin{subfigure}{\linewidth}
                \centering
                \includegraphics[width=1\linewidth]{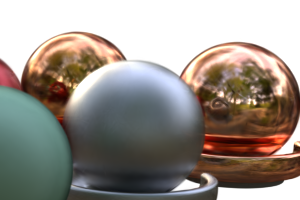}
            \end{subfigure}
            \subcaption{VoD-3DGS[L] (Our method)}   
        \end{subfigure}
    \end{subfigure} 
  \caption{Our method boosts the reflection in the drums (Drums scene), on the hotdog plate (Hotdog scene), and the reflections of the red and gray spheres in the copper materials (Materials scene) of the NeRF-Synthetic \cite{nerf} dataset, achieving state-of-the-art results. }
  \label{fig:full_page_two}
\end{figure*}

As can be seen in Table~\ref{tab:results_real}, our methods achieve state-of-the-art performance on all the datasets used. With the reset on the highest eigenvalue, VoD-3DGS[L] achieves the highest SSIM score in all three and the highest PSNR on the MipNerf360 and the Deep Blending dataset. As for other metrics, VoD-3DGS[L] constantly achieves the top three performance. The findings indicate that the view-dependent opacity can significantly enhance visual quality, with only slightly increased rendering time and memory used. These improvements are particularly noticeable in the specular highlights and varying lighting conditions, as demonstrated in Figures~\ref{fig:full_page_one} and Figure~\ref{fig:full_page_two}.

Our lightweight model, VoD-3DGS[S], also achieves competitive results. Surprisingly, it even outperforms VoD-3DGS[L] on the Tanks \& Temples dataset regarding PSNR. Despite its smaller size, VoD-3DGS[S] consistently ranks among the top three methods, although it falls short in the LPIPS score. As can be seen in Table 1, this method offers the best overall performance and metrics compared to other approaches.

As for the efficiency of our method, even with the trivial implementation, we can achieve real-time results, with $>60$ FPS on average in all datasets, with only a slight increase in memory, significantly outperforming methods like 3DCS \cite{3dcs} when it comes to visual quality and speed. However, efficiency was not the main goal of this paper, and we believe there is significant room for improvement in this regard.

On the Nerf-Synthetic datasets, our methods achieved top-three scores across all statistics, as can be seen in Table~\ref{tab:results_synth}. This time, however, VoD-3DGS[S] outperformed VoD-3DGS[L]. We believe that the smaller size of the synthetic scenes allows VoD-3DGS[S], with its stricter regularization, to identify the optimal number of 3D Gaussians needed to represent the scene without overfitting. Nonetheless, both methods deliver state-of-the-art performance on the Nerf-Synthetic datasets.

\begin{table*}[ht]
\centering
\caption{A qualitative comparison of our method computed over three datasets compared to Spec-Gaussian \cite{spec-gaussian}. Results marked with \textdagger have been recomputed.}
\resizebox{\linewidth}{!}{%
\begin{tabular}{c|ccccccccccccccc|}
\cline{2-16}
                       & \multicolumn{15}{c|}{\textbf{Dataset}}                                                                                \\ \hline
\multicolumn{1}{|c|}{} & \multicolumn{5}{c|}{Mip-NeRF360 Dataset} & \multicolumn{5}{c|}{Tanks \& Temples} & \multicolumn{5}{c|}{Deep Blending} \\
\multicolumn{1}{|c|}{\multirow{-2}{*}{\textbf{Method}}} &
  \textit{LPIPS $\downarrow$} &
  \textit{PSNR $\uparrow$} &
  \textit{SSIM $\uparrow$} &
  FPS &
  \multicolumn{1}{c|}{Mem} &
  \textit{LPIPS $\downarrow$} &
  \textit{PSNR $\uparrow$} &
  \textit{SSIM $\uparrow$} &
  FPS &
  \multicolumn{1}{c|}{Mem} &
  \textit{LPIPS $\downarrow$} &
  \textit{PSNR $\uparrow$} &
  \textit{SSIM $\uparrow$} &
  FPS &
  Mem \\ \midrule
\multicolumn{1}{|c|}{\begin{tabular}[c]{@{}c@{}}Spec Gaussian\\ w/o anchors \textdagger\end{tabular}} &
  N/A &
  N/A &
  N/A &
  N/A &
  \multicolumn{1}{c|}{N/A} &
  \cellcolor[HTML]{F5999A}0.140 &
  23.26 &
  \cellcolor[HTML]{F8CD99}0.855 &
  18 &
  \multicolumn{1}{c|}{1128MB} &
  0.248 &
  27.93 &
  0.885 &
  14 &
  1666MB \\
\multicolumn{1}{|c|}{Spec Gaussian \textdagger} &
  \cellcolor[HTML]{F5999A}0.190 &
  \cellcolor[HTML]{F5999A}27.97 &
  \cellcolor[HTML]{F8CD99}0.817 &
  22 &
  \multicolumn{1}{c|}{776MB} &
  \cellcolor[HTML]{F8CD99}0.156 &
  \cellcolor[HTML]{F5999A}24.53 &
  \cellcolor[HTML]{F5999A}0.860 &
  45 &
  \multicolumn{1}{c|}{247MB} &
  \cellcolor[HTML]{F5999A}0.226 &
  \cellcolor[HTML]{F5999A}30.33 &
  \cellcolor[HTML]{FCF9AD}0.906 &
  130 &
  187MB \\ \hline
\multicolumn{1}{|c|}{\textbf{VoD-GS[S]}} &
  \cellcolor[HTML]{F8CD99}0.217 &
  \cellcolor[HTML]{FCF9AD}27.66 &
  \cellcolor[HTML]{F8CD99}0.817 &
  78 &
  \multicolumn{1}{c|}{687MB} &
  0.165 &
  \cellcolor[HTML]{F8CD99}{\color[HTML]{333333} 24.17} &
  \cellcolor[HTML]{F8CD99}{\color[HTML]{333333} 0.859} &
  106 &
  \multicolumn{1}{c|}{430MB} &
  \cellcolor[HTML]{FCF9AD}0.244 &
  \cellcolor[HTML]{FCF9AD}29.75 &
  \cellcolor[HTML]{F8CD99}0.907 &
  91 &
  559MB \\
\multicolumn{1}{|c|}{\textbf{VoD-GS[L]}} &
  \cellcolor[HTML]{FCF9AD}0.213 &
  \cellcolor[HTML]{F8CD99}{\color[HTML]{333333} 27.79} &
  \cellcolor[HTML]{F5999A}0.818 &
  65 &
  \multicolumn{1}{c|}{849MB} &
  \cellcolor[HTML]{FCF9AD}0.160 &
  \cellcolor[HTML]{F8CD99}23.91 &
  \cellcolor[HTML]{F5999A}0.860 &
  84 &
  \multicolumn{1}{c|}{525MB} &
  \cellcolor[HTML]{F8CD99}0.240 &
  \cellcolor[HTML]{F8CD99}29.84 &
  \cellcolor[HTML]{F5999A}{\color[HTML]{333333} 0.908} &
  85 &
  606MB \\ \bottomrule
\end{tabular}%
}
\label{tab:spec-gaussian}
\end{table*}

\subsection{Results in comparison to Specular Gaussian}
\label{sec:spec}
In this subsection, we evaluate our algorithm in comparison to Spec-Gaussian \cite{spec-gaussian}, a method that utilizes an anisotropic spherical Gaussian (ASG) appearance field with an MLP to model the appearance of each 3D Gaussian. We present our results in Table\ref{tab:spec-gaussian} over the previously mentioned real-life datasets: Mip-Nerf360, Tanks\&Temples and Deep Blending.

We recompute the results from Spec-Gaussian using the same parameters as previously to achieve a fair comparison. However, since Specular Gaussian includes the anchoring technique introduced in the Scaffold-GS \cite{scaffold-gs} and employs a coarse-to-fine strategy, both of which can be effectively integrated into our method, we present two sets of results - with and without these improvements.

As shown in Table~\ref{tab:spec-gaussian}, our performance is slightly lower than that of the full Spec-Gaussian algorithm. However, there is a significant gap between the original method and the extended Spec-Gaussian, which shows a promising future direction of our algorithm. In the case of the MipNerf-360 dataset \cite{mipnerf360}, the Spec-Gaussian version without the anchoring mechanism could not fit into our memory, indicating the potential limitations of the previous work and further showcasing the benefits of both mechanisms present in the full version of Spec-Gaussian that could be incorporated into our work.

\begin{table}[ht]
\caption{An ablation study of our method computed over the Nerf-Synthethic \cite{nerf} dataset.}
\centering
\resizebox{\columnwidth}{!}{%
\begin{tabular}{c|ccc|ccccc|}
\cline{2-9}
\textbf{}                              & \multicolumn{3}{c|}{Adjustments} & \multicolumn{5}{c|}{\textbf{Dataset - Nerf Synthethic}} \\ \hline
\multicolumn{1}{|c|}{\textbf{Method}} &
  Matrix &
  Loss &
  Keypoints &
  LPIPS \textdownarrow &
  PSNR \textuparrow &
  SSIM \textuparrow &
  FPS &
  Memory \\ \midrule
\multicolumn{1}{|l|}{3DGS} & \ding{55}    & \ding{55}    & \ding{55}   & 0.0298     & 33.476     & 0.9700     & 561    & 62MB    \\ \hline
\multicolumn{1}{|c|}{}                 & \ding{51}    & \ding{55}    & \ding{55}   & 0.0303     & 33.513     & 0.9699     & 457    & 64MB    \\
\multicolumn{1}{|c|}{} &
  \ding{51} &
  \ding{51} &
  \ding{55} &
  \cellcolor[HTML]{F5999A}0.0295 &
  \cellcolor[HTML]{F8CD99}33.782 &
  \cellcolor[HTML]{F5999A}0.9711 &
  394 &
  66MB \\
\multicolumn{1}{|c|}{\multirow{-3}{*}{\textbf{VoD-3DGS[S]}}} &
  \ding{51} &
  \ding{51} &
  \ding{51} &
  \cellcolor[HTML]{F8CD99}0.0296 &
  \cellcolor[HTML]{F5999A}33.792 &
  \cellcolor[HTML]{F8CD99}0.9710 &
  396 &
  65MB \\ \hline
\multicolumn{1}{|l|}{\textbf{VoD-3DGS[L]}} &
  \ding{51} &
  \ding{51} &
  \ding{51} &
  \cellcolor[HTML]{FCF9AD}0.0297 &
  \cellcolor[HTML]{FCF9AD}33.690 &
  \cellcolor[HTML]{FCF9AD}0.9706 &
  374 &
  66MB \\ \bottomrule
\end{tabular}%
}

\label{tab:results_ablation_synth}
\end{table}

\subsection{Ablations}

To check the validity of our approach, we conducted an ablation study on the MipNerf-360 dataset presented in Table~\ref{tab:results_ablation_synth}, focusing mostly on the VoD-3DGS[S] approach from our paper. We begin with the standard implementation of 3D Gaussian Splatting, as shown in the first row of the table, and extend it by incorporating the new view-dependent matrix. This involves performing an opacity reset and pruning as described in our method. However, in the second row, we do not regularize the 3D Gaussians based on view consistency. It is evident that even with only this new model, we can significantly enhance the performance compared to the standard 3D Gaussian Splatting.

In the third row of the table, we introduce the proposed view consistency loss. Here, we sample the views used for the view consistency loss computation with uniform probability. Forcing the 3D Gaussians to have a similar opacity when viewed from similar angles slightly increases the number of Gaussians, as evident by the lower FPS count and the higher memory used. However, this change further improves the reported statistics, achieving the highest reported LPIPS and SSIM. 

Finally, in the fourth row, we modify the sampling by selecting views based on the number of pairwise keypoint matches reported by LightGlue. This change shows that, by pairing the cameras with the highest visual content overlap, we can increase the PSNR even further without a significant loss to LPIPS and SSIM, removing between 10000 and 20000  Gaussians and decreasing the memory used. Given the limited scene sizes in the Nerf-Syntehic dataset \cite{nerf}, we believe the decrease will be more noticeable in bigger, real-life scenes like Mip-Nerf360 \cite{mipnerf360}.
\section{Limitations, Conclusion and Future Work}

\subsection{Limitations}
In our paper, we acknowledge some limitations in our method. Specifically, we find that it struggles with scenes that contain less specular content. For instance, in the Treehill or Stump scenes from the MipNerf360 \cite{mipnerf360} dataset, our method yields slightly lower PSNR and SSIM values compared to the original method 3DGS. To address this, we could consider introducing a new regularization term aimed at minimizing the norm of the matrix $\hat{S}_i$. Alternatively, we could explore a different formula for calculating opacity $\hat{\alpha_i}(\omega_j)$ using the matrix $\hat{S}_i$ and the parameter $\gamma_i$.

Additionally, we note a slight decrease in the FPS count and an increase in memory usage. Since efficiency was not the main focus of this paper, we believe there are plenty of potential enhancements from the 3D Gaussian Splatting literature that could be implemented to increase the method's speed and reduce storage requirements. 

Finally, to fully benefit from the view-dependent opacity, our method performs the best when the pair-wise consistency loss is computed with the camera views sampled based on the visual content overlap. In this paper, we implemented the visual content matching naively, but we believe this could be seamlessly integrated with the SfM \cite{sfm} pipeline, required for standard 3D Gaussian Splatting approaches.

\subsection{Conclusion}
The standard 3D Gaussian Splatting faces challenges in accurately representing light interactions that depend on the viewer's perspective. We propose to address this limitation with an enhanced 3D Gaussian Splatting model by introducing a view-dependent opacity. By combining the standard scalar opacity with the square of the projected area of the SGGX \cite{sggx}, computed using a learnable symmetric matrix, we can more effectively capture specular highlights, reflections, and varying lighting conditions.

To ensure a consistent appearance across different views, we regularize the Gaussians through a newly introduced view consistency loss. As a result of these changes, we achieve state-of-the-art qualitative real-time performance across multiple datasets, with only a slight increase in memory usage and a negligible drop in frames per second.

However, our method comes with some limitations. Specifically, we noticed a slight decline in quality for predominantly diffuse scenes. Nevertheless, we believe that our extensions open up a promising avenue for improvement in future work.

\subsection{Future Work}
For future work, exploring other types of extensions to the 3D Gaussian model could lead to even greater improvements in both quality and efficiency. Additionally, with this new extended 3D Gaussian Splatting model, we can not only introduce classical efficiency techniques associated with 3D Gaussians but also develop a new method based on our opacity matrix. Furthermore, by further separating light interaction from the scene surface, we believe a new  3D mesh reconstruction algorithm can be introduced that eliminates 3D Gaussians with highly variable opacity, leading to a higher reconstruction quality of the scene.
\newpage

{
    \small
    \bibliographystyle{ieeenat_fullname}
    \bibliography{references}
}

\end{document}